\documentclass[11pt]{article}
\usepackage{color}
\usepackage{amssymb, graphicx, amsmath}
\usepackage{amsthm}
\usepackage{wrapfig}

\DeclareMathOperator*{\argmin}{arg\,min}
\usepackage{graphicx}
\usepackage{url}

\usepackage{algorithmic}
\usepackage{algorithm}
\usepackage{multirow}
\usepackage{float}
\usepackage[title,toc,titletoc,page]{appendix}
\usepackage{subcaption}

\usepackage{amsfonts,amssymb}
\usepackage{amsmath}
\usepackage{lineno}
\usepackage{flushend}
\usepackage{amssymb}
\usepackage{pifont}

\usepackage{amsfonts,amsmath,amsthm,amsfonts,amssymb}
\usepackage{graphics,epsfig,graphicx}
\usepackage[table]{xcolor}
\usepackage{helvet,epsfig,graphicx}
\usepackage[table]{xcolor}

\usepackage{multirow}

\usepackage{textcomp}

\usepackage{latexsym}
\usepackage{color}
\usepackage{bbm}
\usepackage{amssymb, graphicx, amsmath}
\usepackage{amsthm}
\usepackage{wrapfig}
\usepackage{footnote}
\usepackage{graphicx}
\usepackage{url}
\usepackage{CJK,graphicx}
\usepackage{algorithm}
\usepackage{multirow}
\usepackage{float}
\usepackage{algorithm, algorithmic}
\usepackage{pifont}
\usepackage{amsfonts,amsmath,amsthm,amsfonts,amssymb}
\usepackage{graphics,epsfig,graphicx}
\usepackage[table]{xcolor}
\usepackage{helvet,epsfig,graphicx}

\newtheorem{theorem}{Theorem}[section]

\newtheorem{lemma}[theorem]{Lemma}

\theoremstyle{definition}
\newtheorem{definition}[theorem]{Definition}
\theoremstyle{remark}

\numberwithin{equation}{section}

\newtheorem*{prop*}{Proposition}

\usepackage[colorlinks]{hyperref}

\usepackage{algorithm}
\usepackage{tikz}
\usepackage{amsmath} 
\usepackage{natbib}
\setcitestyle{numbers,square}

\begin{document}

\title{Negotiated Reasoning: On Provably Addressing Relative Over-Generalization}

\author{Junjie~Sheng\thanks{School of Computer Science and Technology, East China Normal University, Shanghai, China. (E-mail: jarvis@stu.ecnu.edu.cn, jwang@cs.ecnu.edu.cn, xfwang@cs.ecnu.edu.cn)}
\and Wenhao~Li\thanks{School of Data Science, The Chinese University of Hong Kong, Shenzhen, Shenzhen Institute of Artificial Intelligence and Robotics for Society, China. (E-mail: \{liwenhao,zhahy\}@cuhk.edu.cn)}
\and Bo~Jin\thanks{School of Software Engineering, Tongji University, Shanghai, China. (E-mail: bjin@tongji.edu.cn)}
\and Hongyuan~Zha$^{\dag}$ 
\and Jun~Wang$^*$
\and Xiangfeng~Wang$^*$
}
\date{}

\maketitle

\begin{abstract}
Over-generalization is a thorny issue in cognitive science, where people may become overly cautious due to past experiences.
Agents in multi-agent reinforcement learning (MARL) also have been found suffering \textit{relative over-generalization} (RO) as people do and stuck to sub-optimal cooperation.
Recent methods have shown that assigning \textit{reasoning} ability to agent can mitigate RO algorithmically and empirically, but there has been a lack of theoretical understanding of RO, let alone designing provably RO-free methods. 
This paper first proves that RO can be avoided when the MARL method satisfies a consistent reasoning requirement under certain conditions. 
Then we introduce a novel reasoning framework, called negotiated reasoning, that first builds the connection between reasoning and RO with theoretical justifications.
After that, we propose an instantiated algorithm, Stein variational negotiated reasoning (SVNR), which uses Stein variational gradient descent to derive a negotiation policy that provably avoids RO in MARL under maximum entropy policy iteration.
The method is further parameterized with neural networks for amortized learning, making computation efficient. 
Numerical experiments on many RO-challenged environments demonstrate the superiority and efficiency of SVNR compared to state-of-the-art methods in addressing RO.
\end{abstract}


\section{Introduction}
\label{sec: intro}
Multi-agent reinforcement learning (MARL) emerged in many areas, such as multiplayer games~\cite{rashid2019starcraft,kurach2020google}, robotics~\cite{ding2020distributed} and traffic controls~\cite{calvo2018heterogeneous}. 
It studies how two or more agents coexisting in a shared environment should act to obtain the highest utility. 
This research paper focuses on the fully cooperative setting, which involves agents working towards the same target of increasing team profitability. 
In particular, we direct attention to the challenge of correcting the pathology of \textit{relative over-generalization} (RO).

Over-generalization is a well-known issue in cognitive science that affects human and animal learning~\cite{rand2014social, laufer2016behavioral, baron2000effects}. 
It occurs when individuals apply broad and often inaccurate rules to specific situations based on limited evidence or experience. 
For instance, the \textit{``once bitten, twice shy''} idiom illustrates this phenomenon: after being bitten by a snake, a man becomes afraid of ropes due to over-generalizing the bitten result with stepping on lines. 
Over-generalization has been observed across various domains such as language acquisition~\cite{gershkoff2006priming}, social learning~\cite{rand2014social}, and decision-making~\cite{laufer2016behavioral}.

Researchers at MARL consider relative over-generalization (RO) to be a significant challenge as it hinders optimal cooperation~\cite{palmer2020independent}. 
Agents learn policies based on their limited interactions and often overfit their actions on the exploration behavior of others, leading to sub-optimal cooperation. For instance, in Particle Gather, where two particles aim to reach a fixed landmark synchronously, each particle is controlled by two agents (\texttt{x}-agent and \texttt{y}-agent) that control horizontal and vertical movement. When agents explore uniformly, they are easy to suffer a loss due to reaching the landmark alone and afraid to reach the landmark after that. 
Consequently, many MARL methods~\cite{maddpg, masql, pr2} get stuck to sub-optimal cooperation where particles avoid reaching the landmark as shown in $\S$\ref{sec: exp}.

In MARL, there are two main ways to mitigate RO: lenient learning and reasoning-endowed methods. Lenient learning methods~\cite{panait2006lenient, wei2016lenient,palmer2017lenient} aim to reduce RO by being lenient on past experiences. They use a temperature factor for each state-action pair to control the amount of leniency, which in turn controls the ignorance of experience towards policy updating and introduces optimism in value function updates. Although some preliminary results have shown success in simple tabular environments, these methods typically require a large number of hyperparameters to be tuned and are not suitable for complex environments. 
Reasoning-endowed methods~\cite{pr2,rommeo,masql} take a further step. They endow agents with reasoning abilities that enable them to better model the behaviors of others instead of changing the weight of past experiences.
For example, \cite{pr2} endows each agent with recursive reasoning ability inspired by human minds~\cite{minds}. During training, each agent models the behavior of others as their best response to its potential behaviors. This enables agents to better account for the behavior of others and empirically mitigates RO. However, there is still a lack of theoretical understanding regarding RO. 
Two questions naturally arise: 

\vspace{5pt}
\centerline{\textit{1) Can relative over-generalization be provably avoided, and if it can?}}
\centerline{\textit{2) How to design a method that provably addresses relative over-generalization?}}
\vspace{5pt}

This paper answers the first question with theoretical justifications and introduces new concepts to analyze Relative Over-generalization (RO) in Multi-Agent Reinforcement Learning (MARL). The current RO is defined on empirical converged joint policy, which makes it difficult to analyze MARL methods before training. To address this issue, we introduce {\em{Perceived Relative Over-generalization}} (PRO) and {\em{Executed Relative Over-generalization}} (ERO), which define RO for each joint policy update and policy execution, respectively. 
The RO is guaranteed to be addressed when ERO is avoided at convergence. With the basis, we prove that RO can be provably avoided when the MARL method satisfies a \textit{consistent reasoning} condition at convergence. 
This condition requires each agent to model the behaviors of others consistently with their updated/executed behaviors.

For the second question, we propose a novel \textit{negotiated reasoning} framework to design MARL methods that satisfy the \textit{consistent reasoning} condition. 
It takes the idea of human beings to reach consistent reasoning through \textit{negotiation}~\cite{kim1996cheap, carnevale1986time} and the graphical model inference through message-passing~\cite{bp}.
Our negotiated reasoning framework allows agents to make explicit reasoning through negotiation policies during training and to make decisions based on the negotiated agreement.
We prove that agents reach consistent reasoning when they reach an agreement on action selection through appropriate negotiation policies.
Then the remaining questions for addressing RO are how to learn negotiation policies and design the algorithm that makes agents reach consistent reasoning at convergence.
We thus propose the Stein Variational Negotiated Reasoning (SVNR).
Concretely, it derives the negotiation policy (how to negotiate) based on Stein variational gradient descent and specifies the negotiation structure (whom to negotiate) as a strict nested structure.
With maximum entropy policy iteration, SVNR is guaranteed to reach consistent reasoning and optimal cooperation at convergence under mild conditions. 

We further parameterize SVNR with neural networks and propose amortized learning to address the intractable computation complexity and inefficient negotiation.
It distills the negotiated reasoning dynamic to guide the neural network updating and approximates many rounds of negotiated reasoning by the single forward of the neural network inference.
Numerical experiments in two challenging environments ({\em{i.e.}}, the \textit{differential games} and \textit{particle world}) show the superiority and efficiency of SVNR in addressing RO compared with the state-of-the-art reasoning methods.

Our contributions are threefold:
1) We confirm the existence of provably addressing RO methods.
2) We propose a novel framework called negotiated reasoning (NR) and specify the Stein variational NR method, which is the first MARL method that can provably address RO. 
3) We propose a practical implementation of SVNR that demonstrates superior performance in achieving global optimal cooperation in RO-challenged tasks.

\section{Related Work}
\paragraph{Approaches for addressing relative over-generalization (RO):} RO has been taken as a critical game pathology in MARL.
There are two main ways to mitigate RO in MARL: 1) lenient learning~\cite{wei2016lenient, panait2006lenient, palmer2017lenient}, and 2) reasoning-endowed methods~\cite{pr2, rommeo, masql}. 
Lenient learning methods propose lenient mechanisms for learning.
\cite{panait2006lenient, wei2016lenient} first propose to be lenient for unsuccessful cooperation and test their methods in matrix games. 
\cite{palmer2017lenient} further extends the idea to multi-agent deep reinforcement learning.
However, there are a lot of hyperparameters to tune, which is impractical. 
Reasoning-endowed methods~\cite{masql, rommeo, pr2} propose to endow reasoning ability for each agent to mitigate RO empirically. 
Specifically, MASQL~\cite{masql} lets each agent models the behavior of others as the induced behaviors of its joint Q function. 
In ROMMEO~\cite{rommeo}, each agent models the behaviors of others as the entropy regularized behaviors.
PR2~\cite{pr2} lets each agent model others as playing the best response to its action selection.
Then each agent updates its policy based on the modeled behaviors of others and these methods empirically mitigate RO in simple tasks (most of them only consider two agents).
However, existing methods mainly study RO on its surface without theoretical justifications. 
Although reasoning has shown the ability to mitigate RO in numerical studies, a theoretical connection between RO and reasoning is lacking. 
There are also a number of heuristic ideas to mitigate RO.
This paper proposes the first reasoning method, Stein variational negotiated reasoning, which provably addresses RO. 

\paragraph{Opponent Modeling:} Our work also has a connection with opponent modeling~\cite{om} (OM), which involves modeling the behavior of others. The traditional OM methods only model an opponent's behavior based on their history, assuming they play stationary policies~\cite{friend, fic}. There are two main limitations to these methods. 
The first one is that these methods tend to work with predefined targets of opponents.
Fictitious play~\cite{fic}, friend-or-foe q~\cite{friend}, and many OM methods~\cite{nashq, correlatedq, minimaxq} make a strong assumption on opponent policies which makes them unsuitable for current MARL where opponents change their policies with learning~\cite{pr2}.
The other limitation is that agents  require the Nash equilibrium to update their Q function during training (e.g., Nash Q learning~\cite{nashq} and Wolf models\cite{wolf}). 
These limitations make it hard to apply traditional OM methods to MARL. 
Compared to the traditional OM methods, our methods do not have these limitations. 
Besides, some popular OM methods have been proposed in these years: reasoning-endowed methods~\cite{pr2, rommeo}, and we have summarized them in the previous subsection.

\paragraph{Probabilistic inference for (MA)RL:} Formulating RL problems as probabilistic inference problems has shown substantial results in obtaining maximum entropy exploration~\cite{sql, sac, rl2pi} and allows a number of inference methods to be adopted. 
These methods embed the problem into a graphical model by modeling the relations among states, actions, next states, and indicators of optimality.  
Then the optimal policy can be recovered by making inferences on the graphical model.
For example, Soft Q-learning~\cite{sql} expresses the optimal policy via a Boltzmann distribution and adopts amortized SVGD~\cite{asvgd} to make approximate sampling on the target distribution. 
Different RL problems, the MARL problem involves a number of agents interacting with each other which makes it non-trivial to make extensions from single agent RL reformulations. 
MASQL~\cite{masql}, ROMMEO~\cite{rommeo}, and PR2~\cite{pr2} let each agent model the relations among states, its actions, the actions of its opponents, next states, and indicators of optimality. 
Each agent expresses the optimal joint policy via a Boltzmann distribution and derives its individual policy and opponent policy accordingly.
However, the opponent policy of the agent is not guaranteed to be consistent with the individual policies of opponents. 
Compared with these methods, the agent in our SVNR perceives opponent policy as consistent with the individual policies of opponents by K-Step negotiation during training. 

\section{Preliminary}

\subsection{Cooperative Stochastic Game}
A Cooperative Stochastic Game (CSG) is commonly used to model cooperation in multi-agent systems~\cite{csg}. 
It is defined by a tuple $(\boldsymbol{\mathcal{S}},\{\boldsymbol{\mathcal{U}}_i\}_{i=1}^N, P, {\mathcal{R}}, \gamma)$, where $N$ is the number of agents;
$\boldsymbol{\mathcal{S}}$ is the state space; 
$\boldsymbol{\mathcal{U}}_i$ represents the action space for agent $i$ with $\boldsymbol{\mathcal{U}}=\times_i\, \boldsymbol{\mathcal{U}}_i$ representing the joint action space; 
$P(\boldsymbol{s}'\mid \boldsymbol{s}, \boldsymbol{u})$ representing the probability that environment transit to $\boldsymbol{s}'$ when taking joint action $\boldsymbol{u}$ at state $\boldsymbol{s}$; 
${\mathcal{R}}:\boldsymbol{\mathcal{S}}\times \boldsymbol{\mathcal{U}} \rightarrow \mathbb{R}$ is the team reward\footnote{The utility, reward and payoff are not distinguished.} function; 
$\gamma\in [0,1]$ is the discount factor. 

The goal for the CSG is to find policies $\{\pi_i\}_{i=1}^N$ that make accumulative reward the highest.
The $\pi_i: \boldsymbol{\mathcal{S}}\rightarrow \boldsymbol{\mathcal{U}}_i$ maps the state to agent $i$'s action and the objective of CSG can be formulated as  
\begin{equation}
    \max_{\pi_1, \dots, \pi_N} \mathcal{E} \left[\sum_{t=1}^\infty \gamma^t \mathcal{R}(\boldsymbol{s}_t, \boldsymbol{u}_t) \right], 
\end{equation}
where $\boldsymbol{u}_t$ is sampled from the policies as $\boldsymbol{u}_t^i\sim \pi_i(\cdot\mid \boldsymbol{s}_t)$. 

\subsection{Multi-Agent Reinforcement Learning}
Multi-agent reinforcement learning (MARL) methods are popular for solving the cooperative stochastic game. 
This paper considers the mainstream of MARL schemes: centralized training decentralized execution (CTDE).
Each agent $i$ holds an execution policy $\bar{\pi}_i(u^i\mid s)$ to make execution in a decentralization manner and a \textit{perceived} joint policy $\hat{\pi}_i(\boldsymbol{u} \mid s)$ to do centralized training.
The \textit{perceived} joint policy can be factorized as  $\hat{\pi}_i = \pi_i \rho_i$, where $\pi_i$ is the individual policy and $\rho_i$ is the perceived opponent policy.
Following MaxEnt MARL~\cite{rommeo, pr2, masql}, each agent $i$ optimizes its policy by minimizing the KL-divergence between its perceived joint policy and the induced optimal joint policy:
\begin{equation}
    \begin{aligned}
        \min_{\pi_i} D_{KL} \left( \hat{\pi}_i\| \pi^*_\alpha \right)
    \end{aligned}
    \label{eq: maxentmarl}
\end{equation}
where $\alpha$ is the factor that controls the relative importance between reward and entropy.
The $\pi^*_\alpha$ is induced by the Boltzmann optimal policy:
\begin{equation}    {\textstyle{\pi^*_\alpha(\boldsymbol{u}\mid \boldsymbol{s}):=\exp \left(\frac{1}{\alpha}\left(Q_{\text {soft }}^*\left(\boldsymbol{s}_t, \boldsymbol{u}_t\right)-V_{\mathrm{soft}}^*\left(\boldsymbol{s}_t\right)\right)\right),}}
    \label{eq: pi*}
\end{equation}
where $Q_{\text {soft }}^*,V_{\mathrm{soft}}^*$ denote optimal, soft state-action and state value function, respectively~\cite{sql}.
After that, each agent $i$ obtains decentralized execution policy as $\bar{\pi}_i(u^i\mid \boldsymbol{s}):= \int \hat{\pi}_id\boldsymbol{u}^{-i}$ and the utility of the decentralized execution is: 
\begin{equation}
    U^{\bar{\pi}}:= \sum_t \mathbb{E}_{(\boldsymbol{s}_t, \boldsymbol{u}_t)\sim\beta_{\bar{\pi}}} \mathcal{R}(\boldsymbol{s}_t, \boldsymbol{u}_t), 
\end{equation}
where $\bar{\pi}: =\prod_i^N \bar{\pi}_i$ is the executed joint policy, and $\beta_{\bar{\pi}}$ is the state-action marginals of the trajectory distribution induced by joint policy $\bar{\pi}$. 

\subsection{Stein Variational Gradient Descent (SVGD)}
SVGD~\cite{stein} is a popular Bayesian inference method that sequentially transforms particles to approximate target distributions. 
Considering a target distribution $p(x)$ where $x\in \mathcal{X}\subset \mathcal{R}^D$, SVGD constructs $q(x)$ from some initial distribution
$$q_0(x):=\frac{1}{M}\sum_{\ell=1}^M \delta_{x^{\ell,0}}(x),$$where $\delta$ is the Dirac delta function, $\{x^{\ell,0}\}_{\ell=1}^M$ are particles at initial, and $M$ is the number of particles. 
Then it transforms particles with transform function $f(x)=x+\epsilon \phi(x)$ where $\epsilon$ is the step size and $\phi:\mathcal{X}\rightarrow \mathcal{R}^D$ is the transform direction. 
To be tractable and flexible, $\phi$ is restricted to a vector-valued reproducing Kernel Hilbert space (RKHS) $\mathcal{H}^D=\mathcal{H}_0\times \cdots \times \mathcal{H}_0$ and $\mathcal{H}_0$ is a scalar-valued RKHS of kernel $k(\cdot, \cdot)$ which is positive definite and in the Stein class of $p$ ({\em{e.g.}}, RBF kernel $k({x}, {y})=\exp (-\|{x}-{y}\|_2^2 /(2 h))$). 
According to Stein theory, the steepest direction that minimizing $D_{KL}(q_f\| p)$ is 
\begin{equation}
    \phi^*({x})=\mathbb{E}_{{y} \sim q}\left[k({x}, {y}) \nabla_{{y}} \log p({y})+\nabla_{{y}} k({x}, {y})\right],
\end{equation}
while $\epsilon$ is small enough. 
Update particles based on $x^{\ell,k}\leftarrow x^{\ell, k-1}+\epsilon \phi^*(x^{\ell,k-1})$ until $\phi^*(x)=0$, SVGD ensures $q=p$ when the iteration ends and $k(x,y)$ is strictly positive definite~\cite{stein}.

MPSVGD~\cite{mpsvgd} is a scalable variant of SVGD that considers the target distribution that can be compactly described by a probabilistic graphical model (PGM).
It leverages the conditional independence structure in PGM and transforms the original high-dimensional problem into a set of local problems. 
Specifically, a PGM $p(x)$ can be factorized as $p(x)\propto \prod_{F\in \mathcal{F}} \psi_F(x_F)$ where the factor $F\subset \{1, \dots, D\}$ is the index set and $x_F=[x_d]_{d\in F}$. Then the Markov blanket for $d$ is
$$\Gamma_d = \left\{ \bigcup\{F: F \ni d\} \right\} \backslash\{d\}$$and it tells the conditional dependence that
$$p(x_d \!\mid\! x_{-d} = p \left( x_d \!\mid\! x_{\Gamma d} \right).$$
MPSVGD updates each dimension $d$ with $T_d: x_d \rightarrow \epsilon \phi_d(x_{S_d})$ where $S_d = \{d\} \cup \Gamma_d$ and $\phi_d\in \mathcal{H}_d$. The $\mathcal{H}_d$ is associated with the local kernel $k_d: X_{S_d}\times X_{S_d} \rightarrow \mathbb{R}$ and 
\begin{equation}
    \phi^*_d({x})=\mathbb{E}_{{y}_{S_d} \sim q}\left[k_d({x}_{S_d}, {y}_{S_d}) \nabla_{{y}_d} \log p({y}_d\mid y_{\Gamma_d})+\nabla_{{y}_d} k_d({x}_{S_d}, {y}_{S_d})\right].
\end{equation}
With enough rounds of updating, the particles converge to the target distribution $p(x)$.

\section{Relative Over-Generalization}

This section defines relative over-generalization (RO) under CTDE MARL contexts. 
Specifically, we propose two concepts, perceived relative over-generalization (PRO) and executed relative over-generalization (ERO), that distinguish different RO in CTDE.
Then, we bridge the two concepts to RO and prove that RO can be avoided when PRO and ERO are addressed under mild conditions.

Relative over-generalization is a critical game pathology in MARL.
It occurs when agents prefer a sub-optimal Nash Equilibrium over an optimal Nash Equilibrium because each agent's individual policy in the sub-optimal equilibrium has a higher utility when paired with arbitrary policies from opponents~\cite{masql}.
This definition assumes MARL methods directly select the joint policy from multiple Nash Equilibriums while these methods make a comparison between the current joint policy and updated joint policy for each updating.
Thus we extend RO by considering each update. 
Besides that, the current CTDE scheme in MARL motivates us to decompose RO to perceived relative over-generalization (PRO) in the training phase and executed relative over-generalization (ERO) in the execution phase.
First, we define ERO, which extends RO at each execution step and identifies whether the optimal cooperation is disturbed due to not knowing the behaviors of opponents.

\begin{definition}[Executed Relative Over-generalization]
Agent $i$ suffers executed relative over-generalization if and only if the utility of executed joint policy can be improved by letting agents know others' actions:
\begin{equation}
    \begin{aligned}
      \max_{{\pi}_i} \left\{ U^{{\pi}_i(u^i\mid s, \boldsymbol{u}^{-i})\prod_{j\neq i}\bar{\pi}^*_j(u^j\mid \boldsymbol{s})} \right\}> U^{\prod_j\bar{\pi}^*_j(u^j\mid \boldsymbol{s})} 
    \end{aligned}
\end{equation}
where ${\pi}_i^* = \arg \min_{\pi_i} D_{KL}(\pi_i\rho_i \|  \pi^*_{\alpha})$ is the $i$'s optimal individual policy with $\rho_i$ and $\bar{\pi}^*_i = \int {\pi}_i^* \rho_i d\boldsymbol{u}^{-i}$ is the executed policy for each agent $i$. 
\end{definition}
It is straightforward that agents do not suffer from RO if all agents are free from ERO at convergence. 

Besides that, agents also suffer from RO during their training phase, and we define Perceived Relative Over-generalization.
\begin{definition}[Perceived Relative Over-generalization]
Agents suffer perceived relative over-generalization if and only if there exists an agent $i$ whose optimal perceived joint policy can be closer to the optimal joint policy when knowing the optimal opponent policy: 
\begin{equation}
\begin{aligned}
    \min_{\pi_i} D_{KL}( \pi_i\rho_i \| \pi^*_\alpha) >  \min_{\pi_i} D_{KL}(\pi_i\pi^*_{\alpha}(\boldsymbol{u}^{-i}) \| \pi^*_\alpha),
\end{aligned}
\end{equation}
where $\pi^*_\alpha$ is the optimal joint policy with entropy factor $\alpha$, and $\pi^*_{\alpha}(\boldsymbol{u}^{-i}):=\int_{u^i} \pi^*_\alpha d u^{i}$ is the optimal opponent policy.
\end{definition}
 
When agents are free from PRO, the perceived optimal joint policy for each agent is equal to the optimal joint policy. 

When each agent $i$ reasons others' behaviors consistent with their optimal policy $\rho_i=\pi_\alpha^*(\boldsymbol{u}^{-i})$ in the training phase, others' exploration will not impact the agent's policy updating and the PRO is avoided. 
If PRO is avoided and $\alpha\rightarrow 0$,  
all agents execute deterministically, the agent's execution will not be impacted by others' exploration stochastic in the execution phase, and ERO is avoided. 
These conditions are denoted as consistent reasoning, and we define them below.
\begin{definition}[Consistent Reasoning]
Agents meet consistent reasoning if and only if all agents reason others' behaviors consistent with their optimal policy $\rho_i=\pi_\alpha^*(\boldsymbol{u}^{-i})$ in the training phase and reason others' behaviors consistent with their executed actions during execution.  
\end{definition}
\begin{figure}[htb!]
    \centering
    \begin{subfigure}[b]{0.25\textwidth}
         \centering
         \includegraphics[width=\textwidth]{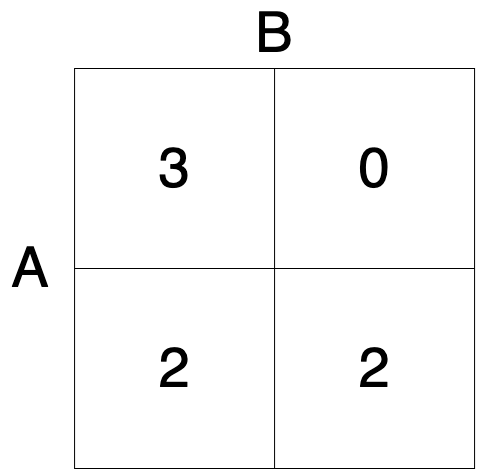}
         \caption{PRO example.}
         \label{fig: payoff}
    \end{subfigure}
    \begin{subfigure}[b]{0.25\textwidth}
         \centering
         \includegraphics[width=\textwidth]{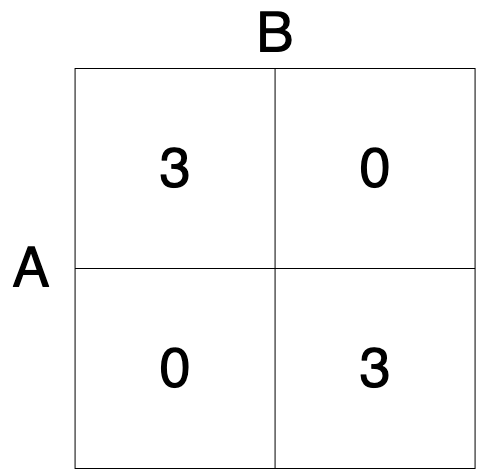}
         \caption{ERO example.}
         \label{fig: payoff1}
    \end{subfigure}
    \caption{The payoff functions.}
    \label{fig: mypayoff}
\end{figure}
When the requirement is met at convergence, agents are free from ERO, and they do not suffer from RO. 
Existing reasoning methods are unable to reach consistent reasoning and suffer from PRO and ERO. 
We take Figure~\ref{fig: mypayoff} as an example to better illustrate how these methods suffer from PRO and ERO. 
It is a single-stage, cooperative game and contains two agents ``A'' and ``B''. 
The action space of each agent is $\{0,1\}$.
In Figrue~\ref{fig: payoff}, MADDPG~\cite{maddpg} usually suffers from PRO due to agents reason others through their historical behaviors.
For agent A, if $\rho_A(0)=\rho_A(1)=0.5$, it will obtain $\hat{\pi}^\prime_A(1,0)=1$ which is sub-optimal. 
MASQL~\cite{masql} usually suffers from ERO in Figure~\ref{fig: payoff1}. 
If $\hat{\pi}^\prime_A(1, 0)=\hat{\pi}^\prime_A(0, 1)=0.5$ and $\hat{\pi}^\prime_B(1, 0)=\hat{\pi}^\prime_B(0, 1)=0.5$, then PRO is avoided. 
However when making decentralized execution based on $\hat{\pi}^\prime$ for each agent, $\bar{\pi}(1,1)=\bar{\pi}(0,0)=\bar{\pi}(1,0)=\bar{\pi}(0,1)=0.25$, which are sub-optimal and suffer from ERO.

\section{Negotiated Reasoning Framework}
Inspired by the critical role of negotiation for consistent reasoning in social cooperation,
we introduce negotiation in the reasoning process to avoid PRO and ERO with theoretical justifications and propose a novel reasoning framework, NR.
In NR, agents take $M$ particles $\{\boldsymbol{u}^{\ell,0}\}_{\ell=1}^M$ to represent the initial perceived joint policy distribution $p(\boldsymbol{u}^0):=\frac{1}{M}\sum_{\ell=1}^M \delta_{\boldsymbol{u}^{\ell,0}}(\boldsymbol{u})$ for a state $s$.
Moreover, each agent $i$ holds a  negotiation ({\em{i.e.}}, perturb) policy $f_i(u_i\mid \boldsymbol{u}_{C_i}, s)$ that updates its action when knowing the $C_i$'s action selection.
Here $C_i\subseteq {1, \dots, N}$ is the negotiated set for agent $i$, which determines whom to negotiate,  $f_i:=\{f_i^1, \dots, f_i^K\}$ where $f_i^k$ is the negotiation policy of agent $i$ in iteration $k$, and $K$ is the number of negotiation rounds which is often large enough.
Then every agent $i$ makes negotiated reasoning as
\begin{equation}
    \begin{aligned}
        u^{\ell,k}_i = f_i^k (u_i\mid \boldsymbol{s}, \boldsymbol{u}_{C_i}^{\ell, k-1}), \quad \forall i\leq N, \ell\leq M, k\leq K.
    \end{aligned}
\end{equation}
Such a negotiation process can be interpreted as agents starting from initial action beliefs and negotiating with each other based on their negotiation policies.
When $f_i^k$ converges to an identity map for each agent, the perceived joint policy converges to a steady perceived joint policy ({\em{i.e.}}, agreement):
$$
\lim_{k \rightarrow K} p(\boldsymbol{u}^k\mid \boldsymbol{s}) := \frac{1}{M}\sum_{\ell=1}^M\delta_{\boldsymbol{u}^{\ell,k}}(\boldsymbol{u})\rightarrow \pi^s(\boldsymbol{u} \mid \boldsymbol{s}), \quad \forall \boldsymbol{u} \in \boldsymbol{\mathcal{U}}.
$$
The negotiated reasoning avoids PRO when it meets certain conditions. 

\begin{theorem}[PRO-free Negotiated Reasoning]
\label{the: pro-free-nr}
For any environment state $s$ where the optimal joint policy is defined as $\pi^*_\alpha$, consider each agent $i$ takes a negotiated reasoning defined on a compact action space $\boldsymbol{\mathcal{U}}_i$, they are PRO-free with $K$ steps negotiated reasoning if 
$$
\lim _{k \rightarrow K} p(\boldsymbol{u}^k \mid \boldsymbol{s}) =\pi^*(\boldsymbol{u}^k\mid \boldsymbol{s}), \quad \forall \boldsymbol{u}^k \in \boldsymbol{\mathcal{U}},
$$
\end{theorem}

This motivates us to learn negotiation policy $f_i$ satisfying the following conditions: 
\begin{equation}
    \begin{aligned}
        &\lim_{k\rightarrow K } f_i^k(u_i\mid \boldsymbol{s}, \boldsymbol{u}_{C_i}^{\ell, k-1}) = u_i^{\ell, k-1},\  \forall\  i\leq N, \ell\leq M , \\
        &\lim_{k \rightarrow K }p(\boldsymbol{u}^k \mid \boldsymbol{s}) = \pi^*(\boldsymbol{u}^k \mid \boldsymbol{s}), \quad \forall \boldsymbol{u}^k \in \boldsymbol{\mathcal{U}}, .
    \end{aligned}
    \label{eq: cond-nego}
\end{equation}
The first condition requires the negotiation policies to converge to the identity map, and the second one requires the perceived joint policy to be identical to the optimal joint policy when the negotiation policy converges. 
We will specify the negotiated policy learning in the following two sections. 

As for ERO-free in decentralized execution, we prove that setting $\bar{\pi}_i=u^{0, K}_i$ with annealing $\alpha \rightarrow 0$ ensures ERO-free in decentralized execution (see proof in Appendix~\ref{proof: ero-free-nr}).
\begin{theorem}[ERO-free Negotiated Reasoning]
For any environment state s, when agents are PRO-free with $K$ reasoning steps, they achieve ERO-free with annealing $\alpha \rightarrow 0$ if each agent $i$ sample action $\bar{\pi}_i=u^{0, K}_i$.
\label{the: ero-free-nr}
\end{theorem}

When all the conditions are met, it is straightforward that consistent reasoning is obtained.
Up to this point, we have established a theoretical connection between reasoning and relative over-generalization (RO). The next step is to design a negotiation policy that satisfies the condition in \eqref{eq: cond-nego} and integrate this negotiated reasoning into existing multi-agent reinforcement learning.

\section{Stein Variational Negotiated Reasoning}
After building the theoretical relationship between reasoning and RO, this section proposes Stein variational NR, SVNR, under the NR framework, which is the first MARL method that provably addresses RO.
We first derive the negotiation policy based on Stein variational gradient descent which obtains PRO-free negotiated reasoning.
Then we devise the policy iteration method of SVNR and prove that it addresses PRO and ERO.
Finally, we propose a practical implementation by parameterizing SVNR with neural networks and amortizing the learning procedure.

\subsection{Learning the negotiation policy}
To learn the negotiation policy that converges to an identity map and lets perceived joint policy converges to the optimal joint policy as in \eqref{eq: cond-nego}, we start by building the relationship between negotiation policy and perceived joint policy. 
Decomposing KL divergence from the perceived joint policy, we have
\begin{equation}
\begin{aligned}
    D_{KL}(p(\boldsymbol{u}\mid \boldsymbol{s})\| \pi^*(\boldsymbol{u}\mid \boldsymbol{s})) = & D_{KL}(p(u_i\mid \boldsymbol{s},\boldsymbol{u}_{-i})p(\boldsymbol{u}_{-i})\| \pi^*(u_i\mid \boldsymbol{s}, \boldsymbol{u}_{-i})p(\boldsymbol{u}_{-i})) \\
    & + D_{KL}(p(\boldsymbol{u}_{-i}\mid \boldsymbol{s})\| \pi^*(\boldsymbol{u}_{-i}\mid \boldsymbol{s})).
\end{aligned}
\end{equation}
It states that the KL divergence between perceived and optimal joint policy can be minimized by
\begin{equation}
    \begin{aligned}
        \min_{p(u_i \mid \boldsymbol{s}, \boldsymbol{u}_{-i})} D_{KL}\big( p({u}_i \!\mid\! \boldsymbol{s},\boldsymbol{u}_{-i})p(\boldsymbol{u}_{-i})\| \pi^*(u_i \mid \boldsymbol{s}, \boldsymbol{u}_{-i})p(\boldsymbol{u}_{-i}) \big),
    \end{aligned}
    \label{eq: dec2}
\end{equation} 
when fixing other agents' action selections (update only one agent's action).
This motivates us to design a negotiation policy that minimizes the \eqref{eq: dec2}. 
One of the most popular ways to solve \eqref{eq: dec2} is (MP)SVGD which can naturally fit the updating of the single agent's action while fixing others'. 
Specifically, it adopts the following scheme, {\em{i.e.}},
\begin{equation}
    f_i(u_i\mid \boldsymbol{u}_{C_i}^{\ell}, \boldsymbol{s}): u_i^{\ell} + \epsilon \phi_i (\boldsymbol{u}_{C_i})^{\ell}, \ \forall \  i\leq N,\ell\leq M, \label{eq: par-ite}
\end{equation} to update the joint policy distribution. 
The $\epsilon$ is the learning rate, and $\phi_i$ is the transformation direction in vector-valued reproducing kernel Hilbert space.
Then the optimal $\phi$ has a closed form solution for \eqref{eq: dec2} when restricting $\|\phi_i\|_{\mathcal{H}_i}\leq1$ and $\epsilon \rightarrow 0$: 
\begin{equation}
    \begin{aligned}
\phi_i^*\left(\boldsymbol{u}_{C_i}\right)= & \mathbb{E}_{\boldsymbol{y}\sim p} \left[k_i\left(\boldsymbol{u}_{C_i}, \boldsymbol{y}_{C_i}\right) \nabla_{y_i} \log \pi^*\left(y_i \mid \boldsymbol{y}_{C_i}\right) + \nabla_{y_i} k_i \left(\boldsymbol{u}_{C_i}, \boldsymbol{y}_{C_i\backslash \{i\}}\right)\right] .
\end{aligned}
\label{eq: phi*}
\end{equation}
The $\phi^*$ provides the steepest direction to optimize the KL divergence.
The Appendix~\ref{der: phi*} shows the details of the derivation.

To further ensure the identity map convergence and let the converged perceived joint policy identical to the optimal joint policy, the design of $\{C_i\}_{i=1}^N$ plays a key role as seen in graphical inference problems~\cite{bp, mpsvgd}. 
Benefiting from the centralized training, we can design $C_i$ without considering communication limitations. 
When $\{C_i\}_{i=1}^N$ satisfies the strict nested requirement ({\em{e.g.}}, $C_i=\{1, \dots, i\}$ for all $i$), negotiated reasoning~\eqref{eq: par-ite} with~\eqref{eq: phi*} converges and the agreement is identical to optimal joint policy ({\em{i.e.}}, satisfies PRO-free conditions \eqref{eq: cond-nego}) as proved in Appendix~\ref{proof: nr-converge}.
We denote the negotiated reasoning with (MP)SVGD and strict nested negotiation set as Stein variational negotiated reasoning.

\subsection{Maximum Entropy Policy Iteration}
It is worth noting that in the previous section, we assumed that the optimal joint policy is known in advance. 
However, agents have to iteratively learn $Q$, and $V$ functions to estimate the optimal joint policy and update their sampling policy accordingly in practice.
This section establishes SVNR on the maximum entropy policy iteration and shows the convergence to the optimal joint policy theoretically.
Concretely, we first define the soft bellman operator as 
\begin{equation}\label{eq: evaluate}
\Gamma_{\hat{\pi}} Q(\boldsymbol{s}_{t}, \boldsymbol{u}_{t}):=r_{t}+\gamma \mathbb{E}_{\boldsymbol{s}_{t+1}}[V(\boldsymbol{s}_{t+1})],
\end{equation}
where $ V(\boldsymbol{s}_{t})=\mathbb{E}_{\hat{\pi}} \left[ Q(\boldsymbol{s}_{t}, \boldsymbol{u}_{t})-\alpha \log \hat{\pi}(\boldsymbol{u}_{t} \mid \boldsymbol{s}_{t}) \right]$.
Each round of iteration usually consists of joint policy evaluation and joint policy improvement, where joint policy evaluation aims to evaluate the policy performance with $Q$ and joint policy improvement updates each agent's policy accordingly.
As for the joint policy evaluation, we obtain the following theorem.
\begin{lemma}[Joint Policy Evaluation]\label{lemma:jointpolicy}
For a mapping $Q^0: \boldsymbol{\mathcal{S}} \times \boldsymbol{\mathcal{U}} \rightarrow \mathbb{R}$ with $|\boldsymbol{\mathcal{U}}| < \infty$, define the $Q^{k+1}=\Gamma_{\hat{\pi}}Q^k$ where the $\Gamma$ is the soft bellman operator, then it converges to the joint soft $Q$-function of $\hat{\pi}$ as $k\rightarrow \infty$.
\end{lemma}

Following \eqref{eq: pi*} and \eqref{eq: phi*}, the $\hat{\pi}$ is updated as:
\begin{equation}
    \begin{aligned}
    & \hat{\pi}(\boldsymbol{u}) = \lim_{k\rightarrow K}\frac{1}{M} \sum_{\ell=1}^M \delta_{u^{\ell, k}} (\boldsymbol{u}), \\
    & u^{\ell, k}_i = u^{\ell, k-1} + \epsilon \phi^*(\boldsymbol{u}^{\ell, k}_{C_i}, u_i^{\ell, k-1}), \ \forall i\leq N, \ell\leq M, k\leq K,\\
    & \Tilde{\pi} = \exp{\frac{1}{\alpha} \left( Q(u_i, \boldsymbol{u}_{C_i}, \boldsymbol{s})-V(\boldsymbol{u}_{C_i},\boldsymbol{s}) \right)},
    \end{aligned}
    \label{eq: new-pi}
\end{equation}
where
$$Q(u_i, \boldsymbol{u}_{C_i}, \boldsymbol{s}) = \mathbb{E}_{\bar{\boldsymbol{u}}\sim \hat{\pi}(\boldsymbol{s}), \bar{\boldsymbol{u}}_{C_i}=\boldsymbol{u}_{C_i}, \bar{\boldsymbol{u}}_i=\boldsymbol{u}_{i}} Q( \bar{\boldsymbol{u}}, \boldsymbol{s}), \quad V_i(\boldsymbol{u}_{C_i}, \boldsymbol{s}) = \mathbb{E}_{\bar{\boldsymbol{u}}^\prime \sim \hat{\pi}(\boldsymbol{s}),\bar{\boldsymbol{u}}_{C_i}=\boldsymbol{u}_{C_i} } Q(\bar{\boldsymbol{u}}, \boldsymbol{s}),$$and $\phi^*_i$ take $\Tilde{\pi}$ instead of $\pi^*$ to construct the SVGD direction.
Then we can obtain the following joint policy improvement lemma:
\begin{lemma}[Policy Improvement]\label{lemma:coorpolicy}
When the negotiation policies satisfy the strict nested requirement, given the current perceived joint policy as $\hat{\pi}$, 
update it based on the \eqref{eq: new-pi} and obtain the new perceived joint policy $\hat{\pi}^\prime$.
The $Q^{\hat{\pi}^\prime}(\boldsymbol{s}_t, \boldsymbol{u}_t) \geq Q^{\hat{\pi}}(\boldsymbol{s}_t, \boldsymbol{u}_t)$ with $|\boldsymbol{\mathcal{U}}| < \infty$.
\end{lemma}
Following Lemma \ref{lemma:jointpolicy} and Lemma \ref{lemma:coorpolicy}, we can establish the following SVNR policy iteration theorem and our proposed coordinated policy iteration method accordingly.
\begin{theorem}[SVNR Policy Iteration]\label{theorem:coorpolicy}
When the individual policies satisfy the strict nested  requirement,
considering repeated apply the joint policy evaluation and joint policy improvement on the perceived joint policy $\hat{\pi}$, then $\hat{\pi}$ will converge to ${\pi}^{*}$ that makes
$$Q^{{\pi}^{*}} \left( \boldsymbol{s}_t, \boldsymbol{u}_t \right) \geq Q^{\hat{\pi}} \left( \boldsymbol{s}_t, \boldsymbol{u}_t \right),\quad \forall \hat{\pi} \in \Pi,\  \left( \boldsymbol{s}_t, \boldsymbol{u}_t \right) \in \boldsymbol{\mathcal{S}}\times \boldsymbol{\mathcal{U}},\  \big| \boldsymbol{\mathcal{U}} \big| < \infty.$$
\end{theorem}
Based on the Theorem~\ref{theorem:coorpolicy}, we can obtain the convergence of SVNR policy iteration to the optimal joint policy. 
Further, taking Theorem~\ref{the: ero-free-nr}, we can obtain ERO-free executed joint policy $\bar{\pi}$ by annealing $\alpha$ to a small enough number.

However, empirically, the SVNR policy iteration assumes knowing the word model and encounters high computation and storage complexity due to 
1) {\em{inefficient policy representation}}: SVNR policy iteration represents the joint policy with particles that scale poorly on state-action space;
2) {\em{intractable optimization}}: During learning, the soft bellman operator takes expectations on both the state distribution and joint policy distribution, which is intractable in realistic settings.
To this end, we propose a practical implementation for SVNR.

\section{A Practical Implementation of SVNR}
\begin{figure*}[h!]
    \centering
    \includegraphics[width=0.85\textwidth]{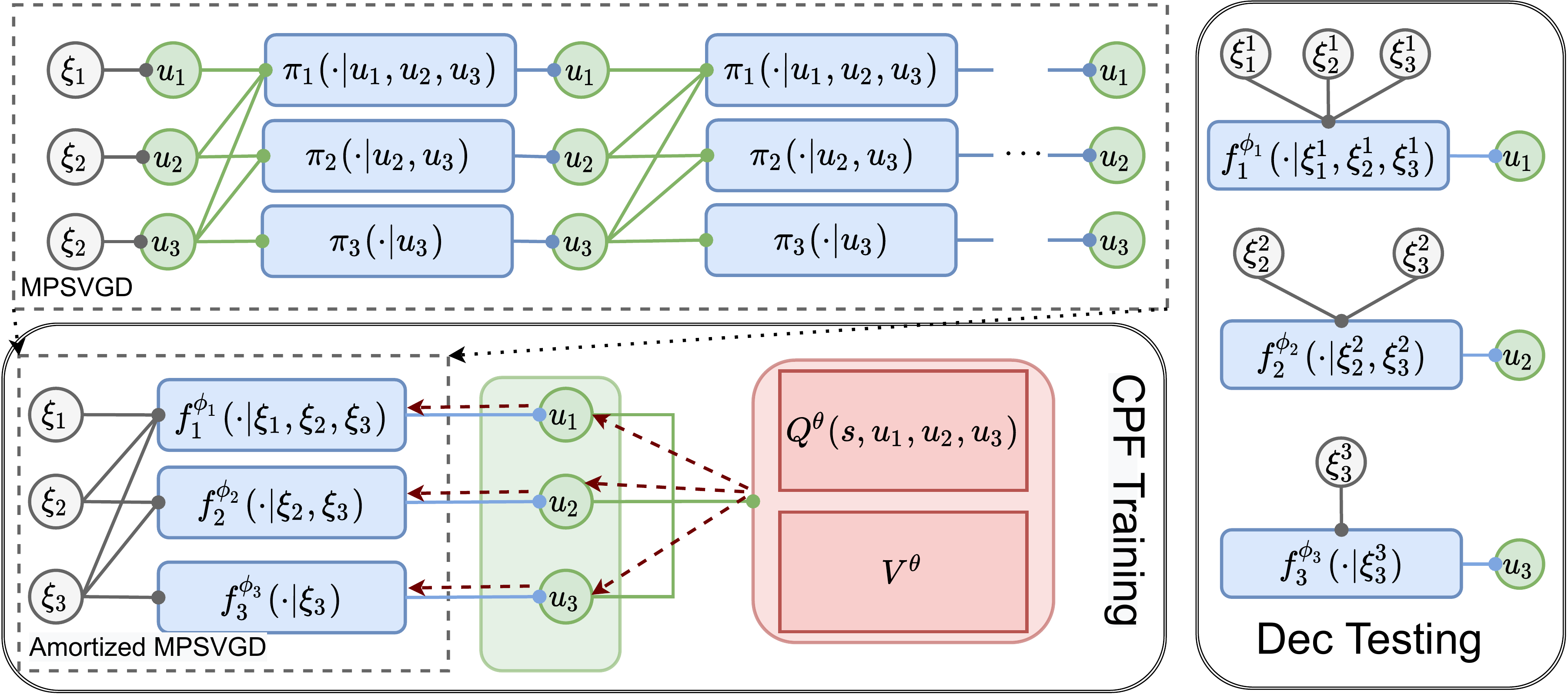}
    \caption{The practical Stein Variational Negotiated Reasoning (SVNR) in the $3$-agents system. SVNR adopts nested negotiated reasoning and adopts amortized MPSVGD to output the actions. The amortized MPSVGD  distills the multi-rounds negotiated reasoning dynamic by well-established neural networks. The ``Dec Testing'' (rightmost part) illustrates how the proposed SVNR executes in a decentralized manner.}
    \label{fig: SVNR}
\end{figure*}
To address the inefficient policy representation and intractable optimization issues, this section adopts neural networks to parameterize the policies and learn them with the proposed amortized optimization.
To gain efficient action sampling, we propose Amortized MPSVGD.
It aims to adopt neural networks to distill the SVNR's update dynamic and approximate the steady negotiation result in $\hat{\pi}(\boldsymbol{u})$ with the neural network inference.
Formally, each agent holds a stochastic mapping function $u_i=f^{\psi_i}_i(\cdot | \xi_i, \boldsymbol{\xi}_{C_i}, s)$ that maps initial noises ({\em{i.e.}}, gaussian noises) to its action distribution.
The $\xi_i$ is the noise drawn by agent $i$. 
We denote the induced joint distribution as
$$p^{\boldsymbol{\psi}} \left( \boldsymbol{u}\!\mid\! \boldsymbol{s}, \boldsymbol{\xi} \right) := \prod_{i=1}^N f_i^{\psi_i} \left( u_i \!\mid\! \xi_i, \boldsymbol{\xi}_{C_i}, \boldsymbol{s} \right).$$
The goal of the proposed amortized MPSVGD method is to find $\boldsymbol{\psi}^*$ that satisfies:
\begin{equation}
\argmin_{\hat{\boldsymbol{\psi}}}\ \mathrm{KL} \left( p^{\hat{\boldsymbol{\psi}}} \left( \cdot \mid \boldsymbol{s}, \boldsymbol{\xi} \right)\, \Vert\, \hat{\pi} \left( \boldsymbol{u} \right) \right).
\label{eq: obj}
\end{equation}
A straightforward way to learn $\psi$ is to iterate the \eqref{eq: new-pi} procedure until convergence and to establish the neural networks $\{ \psi_1, \dots, \psi_N \}$ which can fit the agreement.
However, the \eqref{eq: new-pi} requires many rounds of updating, and this motivates us to introduce an incremental update scheme.
For each agent $i$, its policy parameter $\psi_i$ is updated by moving along its SVGD's gradient in order to approach the target joint policy.
Sampling joint actions $\boldsymbol{u}^1, \dots, \boldsymbol{u}^M$ from $p$ and assuming we can perturb agent $i$'s action $u_i^j = f^{\psi_i}(\xi_i^j; \xi_{C_i}^j, \boldsymbol{s})$ in appropriate direction $\Delta f^{\psi_i}(\xi_{i}^j ;\xi_{C_i}^j, \boldsymbol{s})$, the induced KL divergence in \eqref{eq: obj} can further be reduced. MPSVGD provides the most greedy direction as
\begin{equation}\label{eq: ACF-grad0}
\Delta f^{\boldsymbol{\psi}}_i (\cdot ; \boldsymbol{s}_{t}) =\mathbb{E}_{\boldsymbol{u}\sim p^{\psi}} \left[ \kappa_i \big(\boldsymbol{u}_{C_i}, p^{\boldsymbol{\psi}}_{C_i}(\cdot ; {s}_{t})\big)\nabla_{u^{\prime}_i} Q^{\theta}({\boldsymbol{s}}_{t}, \boldsymbol{u}^{\prime})\big|_{\boldsymbol{u}^{\prime}=\boldsymbol{u}} + \alpha_i \nabla_{\boldsymbol{u}^{\prime}_i} \kappa_i \big(\boldsymbol{u}_{C_i}^{\prime}, p^{\boldsymbol{\psi}}_{C_i}(\cdot ; {\boldsymbol{s}}_{t}) \big) \big|_{\boldsymbol{u}^{\prime}=\boldsymbol{u}} \right],
\end{equation}
where $\alpha_i$ is the agent $i$'s temperature term, $\theta$ is the neural network paramter of central critic, and $\kappa_i$ is the agent $i$'s kernel function as in MPSVGD.
As explained in \cite{asvgd}, we can set $\frac{\partial J_{p}(\boldsymbol{\phi}; \boldsymbol{s}_t)}{\partial u_i} \propto \Delta f^{\boldsymbol{\phi}}_i$.
Further, the gradient in MPSVGD can be backpropagated to the mapping network $\phi_i$, {\em{i.e.}},
\begin{equation}
    {\textstyle{\frac{\partial J_{p}(\boldsymbol{\psi} ; \boldsymbol{s}_{t})}{\partial \psi_i} \propto \mathbb{E}_{\xi} \left[\Delta f^{\psi}_i (\xi ; \boldsymbol{s}_{t}) \frac{\partial f^{\psi}_i(\xi; \boldsymbol{s}_{t})}{\partial \psi_i}\right].}}
    \label{eq: ACF-grad2}
\end{equation}
Therefore, any gradient-based optimization methods can optimize the parameters $\psi_i$.
The detailed derivations of \eqref{eq: ACF-grad0} and
\eqref{eq: ACF-grad2} are shown in Appendix~\ref{sec: der}.
With this Amortized MPSVGD mapping function, neural network inference can directly sample joint actions.

Furthermore, we consider the intractable evaluation step as in \eqref{eq: evaluate}.
Inspired by soft $Q$-learning~\cite{sql}, we can transform the fixed point iteration to the stochastic optimization on minimizing the $\|\Gamma_Q - Q\|$.
Specifically, the importance sampling is adopted to approximate the value function and minimize the bellman error:
\begin{equation}\label{eq: critic-upd}
\begin{aligned}
{\textstyle{\theta^{\mathrm{new}}=\argmin_{\theta^\prime} \mathbb{E} \left[\frac{1}{2}(r + V^{{\theta}}({\boldsymbol{s}}_{t+1})-Q^{\theta^\prime}({\boldsymbol{s}}_{t}, \boldsymbol{u}))^{2} \right],}}
\end{aligned}
\end{equation}
where the expectation is taken on ${\boldsymbol{s}}_{t}, \boldsymbol{u}, r, {s}_{t+1} \sim D$ and
$${\textstyle{V^{{{\theta}}}\left( {\boldsymbol{s}}_{t} \right) :=  \alpha \log \mathbb{E}_{{\boldsymbol{u}}^{\prime}\sim p(\cdot\mid \boldsymbol{s}_t)} \left[ \exp \left( \frac{1}{\alpha} Q^{{{\theta}}} \left( {\boldsymbol{s}}_{t}, {\boldsymbol{u}}^{\prime} \right) \right) \right]}}.$$ 
We summarize the proposed {\bf{SVNR}} in Figure~\ref{fig: SVNR}, with pseudocode in Algorithm~\ref{alg: SVNR}.
It adopts amortized MPSVGD with a centralized critic to learn the policy for each agent.
Each agent $i$ holds its conditional policy $f^{\psi_i}(a_i \mid a_{C_i}, \boldsymbol{s})$ with $\{C_i\}_{i=1}^N$ as strict nested set.
In the execution stage, all agents will initialize their actions randomly as $\{ \xi_i \}$ while each $\xi_i$ will be shared with its neighbors $C_i$.
The action of agent $i$ is generated by $f^{\psi_i}(\xi_i; \xi_{C_i}, \boldsymbol{s})$ based on received noises $(\xi_i, \xi_{C_i})$ and state $\boldsymbol{s}$.
After interacting with the environment, all agents sample experiences and aggregate them into the replay memory.
Further, based on \eqref{eq: ACF-grad0} and \eqref{eq: critic-upd}, each agent's policy can be updated in the learning phase.

\begin{algorithm}[tb]
\caption{{\bf{SVNR}}: Stein Variational  Negotiated Reasoning Method}\label{alg: SVNR}
\begin{algorithmic}
\STATE {\bfseries Input:}{Initial policy $f^{\psi_i}$ for every agent $i$; centralized critic $Q^{\theta}$; coordination edges $\mathbf{C}$; empty replay buffer $\mathcal{D}$; kernel function $\kappa_i$ for agent $i$; particle numbers $K$; target critic as $Q^{\bar{\theta}}:= Q^{\theta}$.}

\WHILE{not convergence}
\STATE \textbf{Collect Experiences:}

\STATE \quad Each agent $i$ samples noise: $\xi_i\in\mathcal{N}(0, I)$;

\STATE \quad Each agent $i$ samples actions for state $\boldsymbol{s}$, {\em{i.e.}}, $u_i\leftarrow f^{\psi_i} (\xi_i; \xi_{C_i}, \boldsymbol{s});$

\STATE \quad Execute the joint action $a:=\{a_1, \dots, a_N\}$ in the environment;

\STATE \quad Observe the next state $\boldsymbol{s}^\prime$, reward $r$;

\STATE \quad Add new experiences into the replay buffer, {\em{i.e.}}, $\mathcal{D}\leftarrow \mathcal{D} \cup \left\{(\boldsymbol{s}, \boldsymbol{u}, r, \boldsymbol{s}^\prime) \right\}.$

\STATE \textbf{Sample Experiences:}

\STATE \quad Sample a mini-batch from the replay memory, i.e,
$\{(\boldsymbol{s}, \boldsymbol{u}, r, \boldsymbol{s}^\prime), \dots\} \sim \mathcal{D}.$

\STATE \textbf{Update Value Functions:}

\STATE \quad For each agent $i$, sample $\{u_i^{\ell}\}_{\ell=1}^M$ for state $\boldsymbol{s}^\prime$;

\STATE \quad Update $\theta$ based on equation \eqref{eq: critic-upd}.

\STATE \textbf{Update Policies:}

\STATE \quad Sample $k$ noise signals for agent $i$ at state $\boldsymbol{s}$, {\em{i.e.}},
$\xi_i^{\ell} \in \mathcal{N}(0,I),\quad \forall \ell=1,\cdots,M;$

\STATE \quad Generate $k$ joint actions for state $\boldsymbol{s}_t$, {\em{i.e.}}, $u_i^{\ell}\leftarrow f^{\psi_i}(\xi_i^{\ell}; \boldsymbol{\xi}_{C_i}^{\ell}, \boldsymbol{s}), \quad \forall \ell=1,\cdots,M;$

\STATE \quad Calculate $\Delta f^{{\psi}_i}$ based on \eqref{eq: ACF-grad0} for each agent $i$;

\STATE \quad Calculate the gradient of $\psi_i$ by \eqref{eq: ACF-grad2} and update $\psi_i$ using ADAM.

\IF {time to update}
\STATE Update target parameters: $\bar{\theta} \rightarrow \theta$.
\ENDIF 
\ENDWHILE
\end{algorithmic}
\end{algorithm}

\section{Experiments}
\label{sec: exp}
\begin{table*}[htb!]
\caption{Execution performances in testings. The proposed SVNR achieves the highest returns in all scenarios.}
\centering
\begin{tabular}{|l|l|l|l|l|}
\hline
Methods / Scenarios & \begin{tabular}[c]{@{}l@{}}\textit{Max Of Three}\\ ($s_2=3.0$)\end{tabular} & \begin{tabular}[c]{@{}l@{}}\textit{Max Of Three}\\ ($s_2=2.0$)\end{tabular} & \begin{tabular}[c]{@{}l@{}}\textit{Max Of Three}\\ ($s_2=1.5$)\end{tabular} & \textit{Particle Gather} \\ \hline
SVNR~(Ours)          & $\mathbf{9.60 \pm 0.30}$                                      & $\mathbf{9.64 \pm 0.17}$                                       & $\mathbf{9.71 \pm 0.20}$                                                  & $\mathbf{4.76 \pm 0.20}$  \\ \hline
MADDPG       & $2.08 \pm 4.63$                                                 & $-0.66 \pm 0.67$                                                 & $-0.64 \pm 0.43$                                              & $0.00 \pm 0.00$  \\ \hline
MASQL        & $8.92 \pm 0.37$                                                    & $-0.58 \pm 0.24$                                                  & $-0.34 \pm 0.12$                                                   &  $-0.54 \pm 0.20$ \\ \hline
PR2          & $4.76 \pm 3.64$ & $-0.64 \pm 0.45$ & $-0.29 \pm 0.10$                                                  & $0.00\pm0.02$   \\ \hline
ROMMEO       & $6.14 \pm 4.82$ & $1.59 \pm 5.03$ & $-0.59 \pm 0.25$                                                   &  $-0.87 \pm 0.22$ \\ \hline
L-MADRL       & $9.54 \pm 0.13$ & $1.63 \pm 2.51$ & $-0.07 \pm 0.04$                                                &  $-0.75 \pm 0.00$\\ \hline
\end{tabular}
\label{tb: dec-result}
\end{table*}

To evaluate the performance of SVNR in RO-challenged MARL problems, we take two differential games (\textit{Two Modalities} and \textit{Max of Three}~\cite{differential}) and the \textit{Particle Gather}~\cite{mordatch2018emergence}) as our testbeds.
They pose strong RO challenges for MARL methods and are taken as popular testbeds for performance evaluation on addressing RO. 


\paragraph{Baselines.} The baselines include the popular MARL methods and methods that aim to address RO, {\em{i.e.}}, MADDPG~\cite{maddpg}, MASQL~\cite{masql}, PR2~\cite{pr2}, ROMMEO~\cite{rommeo}, and Lenient MADRL~\cite{palmer2017lenient}.

\paragraph{Hyperparameters.} For SVNR, we take the negotiation set: $C_i=\{1, \dots, i\},\  \forall  i$.
For all experiments, we use the TPE Sampler~\cite{tpe} to select the learning rates, particle numbers, and the entropy coefficient $\alpha$ based on the maximum mean reward in $50$ trails.
The learning rate and initial $\alpha$ are finetuned in $[10^{-4}, 10^{-1}]$ and $[10^{-1}, 10]$, and particle numbers are finetuned in an integer space from $16$ to $64$.
Other hyperparameters follow the ROMMEO\footnote{\url{https://github.com/rommeoijcai2019/rommeo}}.
The optimizer is ADAM, and the sizes of the replay buffer and batch are $10^6$ and $512$. 
$k(x, x^{\prime})=\exp (-1/h\|x-x^{\prime}\|_{2}^{2})$, bandwidth $h=\operatorname{med}^{2} / \log n$, where $\operatorname{med}$ is the median of the pairwise distance between the current points $\{x_{i}\}_{i=1}^{n}$  as suggested in amortized SVGD~\cite{asvgd}.
To gain exploration in the early stage, we anneal $\alpha$ based on $\alpha=\alpha^\prime + \exp(-0.1 \times \max(\mathrm{steps}-10, 0))*500$ all methods in most of the scenarios where $\alpha^\prime$ is the initial $\alpha$.
The only exception is that we anneal $\alpha$ to $1$ when we investigate the PRO issue for all the methods.

\subsection{The Differential Game}\label{sec:diff-game}
The differential game is a flexible and wide-adopted framework to design a challenging stateless MARL environment.
We consider a three-agents case.
Each agent shares a common one-dimension bounded continuous action space of $[-10, 10]$.
Their rewards are shared and determined by their joint action under the reward function
$r(u_{1}, u_{2}, u_{3})=\max (g_{1}, g_{2}),$
where $a_1, a_2, a_3$ are actions of $3$ agents respectively, and
\begin{equation}
\begin{aligned}
    {\textstyle{g_{1}=0.8 \times \left[-\big(\frac{u_{1}+5}{3}\big)^{2}-\big(\frac{u_{2}+5}{3}\big)^{2} - \big(\frac{u_{3}-3}{3}\big)^{2} \right]+c_1,}}\\
    {\textstyle{g_{2}= h_2\times \left[- \big(\frac{u_{1}-x_2}{s_{2}}\big)^{2}-\big(\frac{u_{2}-y_2}{s_{2}}\big)^{2}-\big(\frac{u_{3}-z_2}{s_{2}}\big)^{2} \right]+c_2.}}
\end{aligned}
\nonumber
\end{equation}
This is a flexible environment to evaluate RO game pathology in MARL.
The $s_2$ controls the coverage of optimal solution, and setting it with a high value makes MARL methods easy to suffer from ERO.
Thus we devise the \textit{Max Of Three} scenario as the ERO-Challenged scenario, which sets $s_2$ to different values, to evaluate the performance on addressing ERO. 
Setting $c_1=c_2$ results in two-modality, which raises the difficulty for agents to obtain the optimal perceived joint policy and thus is a PRO-challenged environment.
We thus devise the \textit{Two Modalities } scenario as the PRO-Challenged scenario accordingly.

\begin{figure}[htb!]
    \centering
    \begin{subfigure}[b]{0.32\textwidth}
         \centering
         \includegraphics[width=\textwidth]{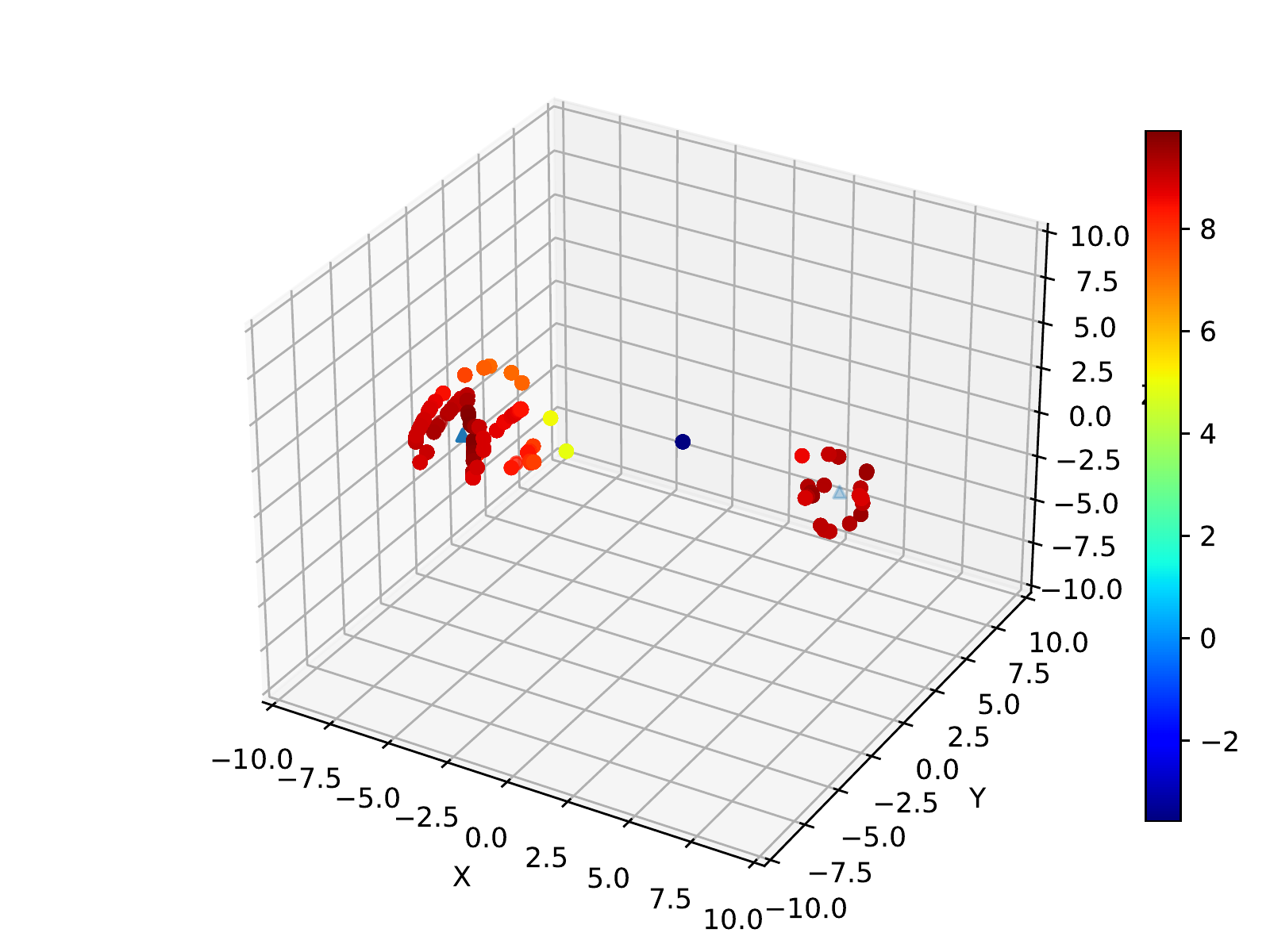}
         \caption{SVNR~(Ours)}
         \label{fig: goal-GPF}
    \end{subfigure}
    \begin{subfigure}[b]{0.32\textwidth}
         \centering
         \includegraphics[width=\textwidth]{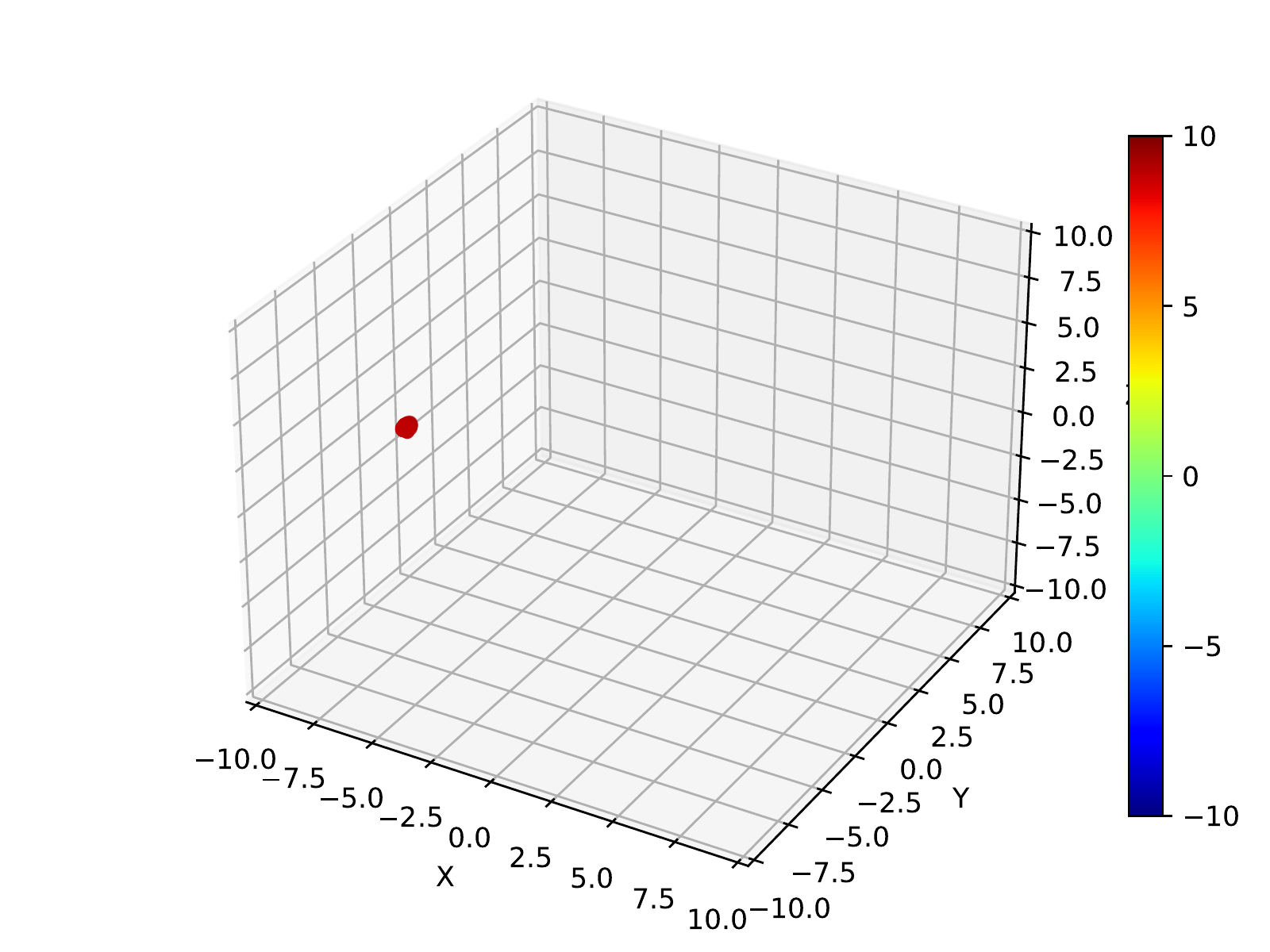}
         \caption{MADDPG}
    \end{subfigure}
    \begin{subfigure}[b]{0.32\textwidth}
         \centering
         \includegraphics[width=\textwidth]{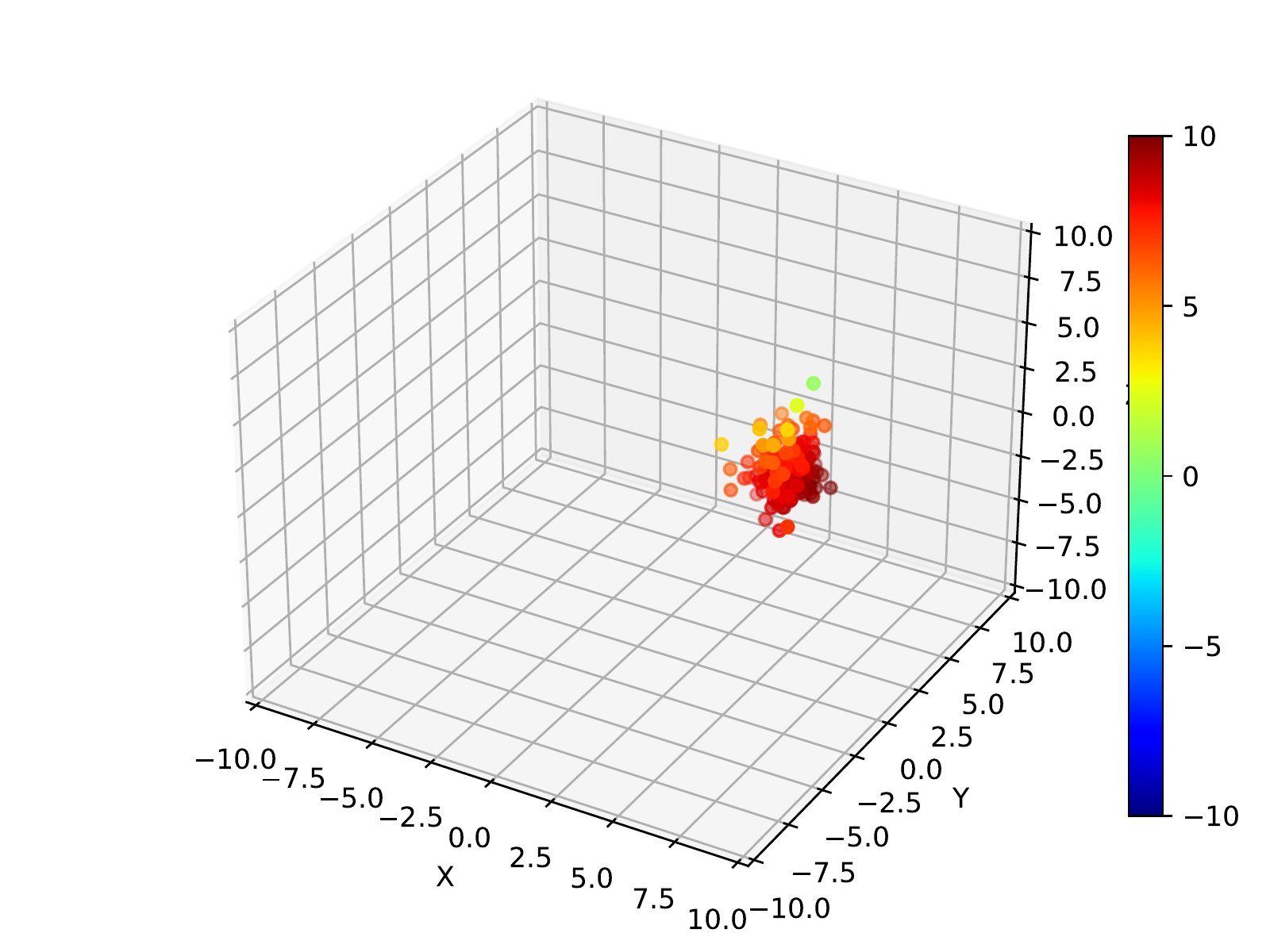}
         \caption{MASQL}
    \end{subfigure}
    
    \begin{subfigure}[b]{0.32\textwidth}
         \centering
         \includegraphics[width=\textwidth]{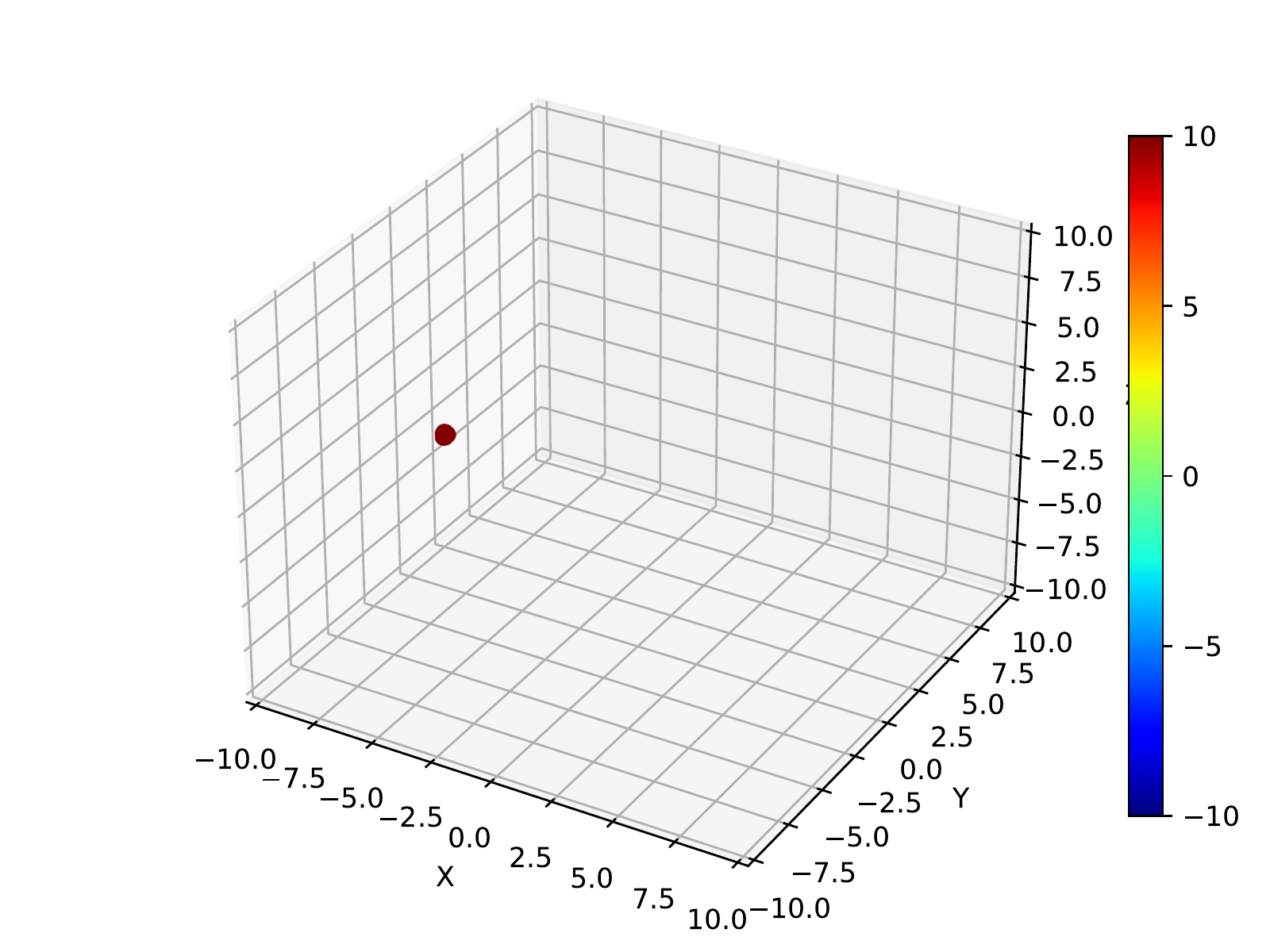}
         \caption{PR2}
    \end{subfigure}
    \begin{subfigure}[b]{0.32\textwidth}
         \centering
         \includegraphics[width=\textwidth]{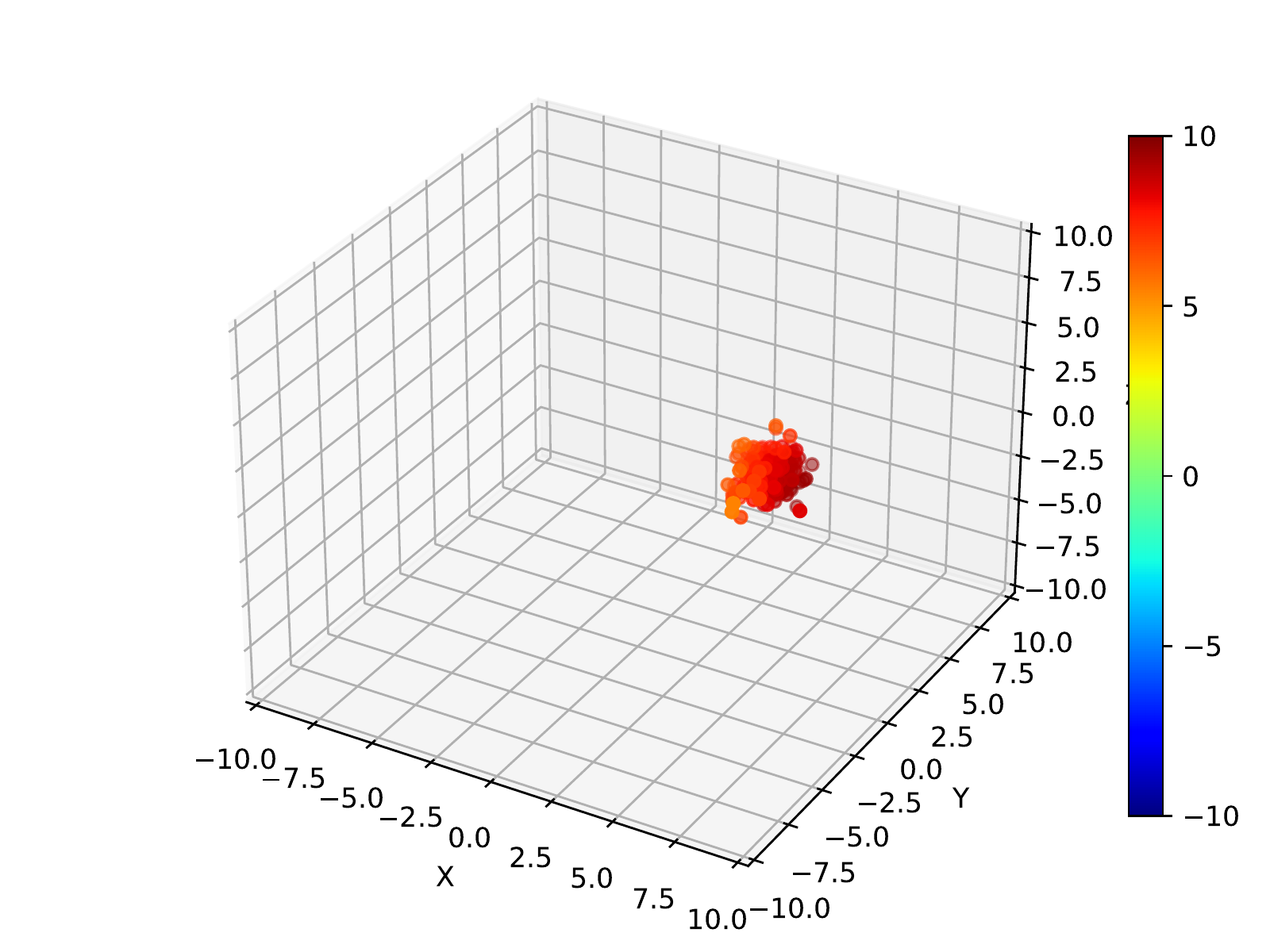}
         \caption{ROMMEO}
         \label{fig: goal-rommeo}
    \end{subfigure}
    \begin{subfigure}[b]{0.32\textwidth}
         \centering
         \includegraphics[width=\textwidth]{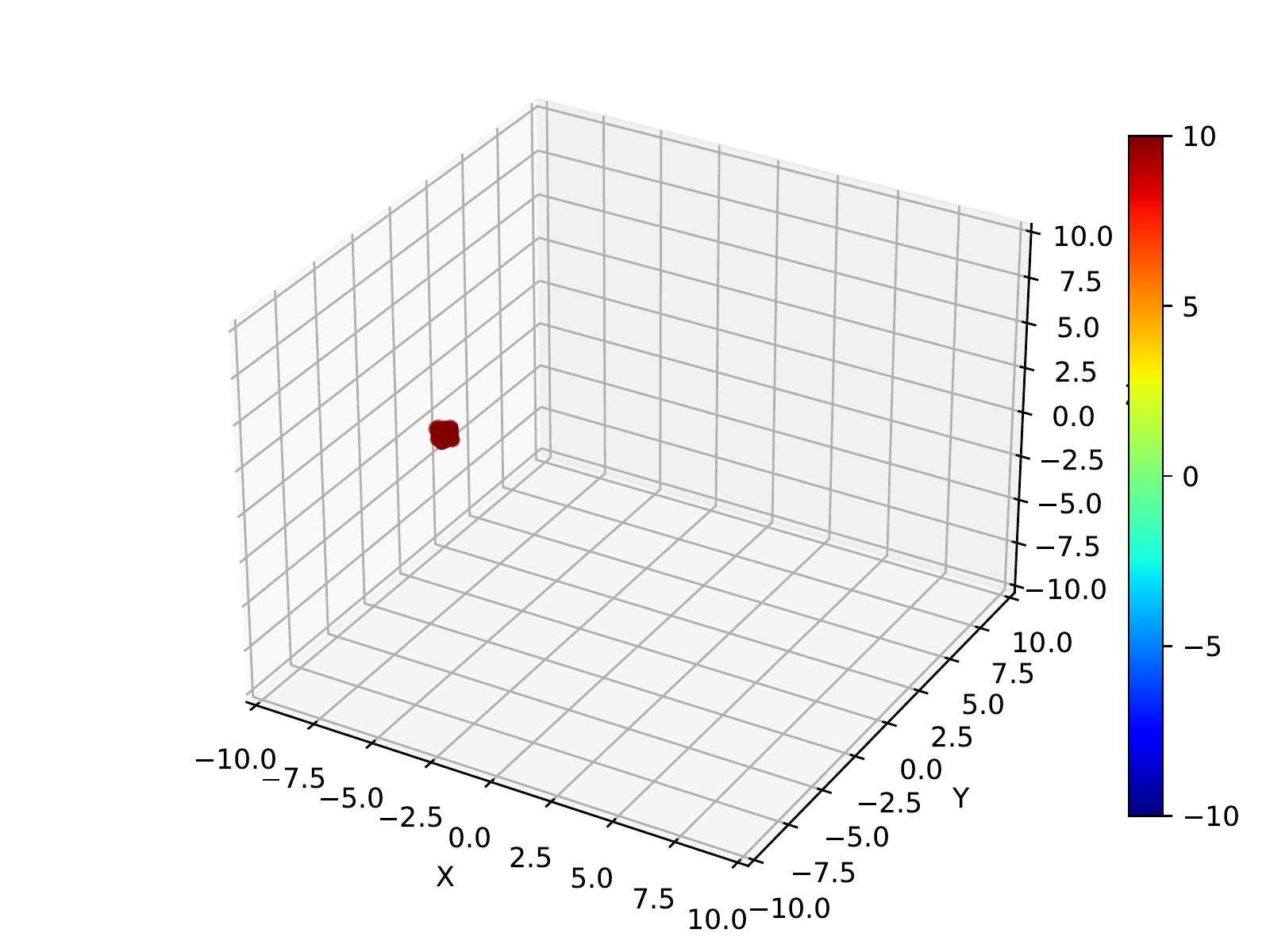}
         \caption{L-MADRL}
         \label{fig: goal-lenient}
    \end{subfigure}
    \caption{The converged perceived joint policy visualization in \textit{Two Modalities} scenario. The optimal perceived joint policy should capture both modalities, and only our SVNR captures the two modalities.}
    \label{fig: two-modality}
\end{figure}

\paragraph{ERO-Challenged.} We consider a difficult scenario for continuous MARL, \textit{Max of Three}, which is extended from the \textit{Max of Two}~\cite{rommeo, masql, pr2}.
Specifically, we set the $h_2=1, x_2=7, y_2=7, z_2=-4, c_1=0, c_2=10$.
By setting different values for $s_2$, we can flexibly control how the RO issue affects the agents.
The smaller the $s_2$, the smaller the coverage of $g_2$, and the more severe the RO issue.
We examine different methods under different $s_2$, {\em{i.e.}}, $s_2=1.5$, $s_2=2.0$ and $s_2=3.0$ and $5000$ episodes are used for all cases.

\begin{figure*}[htb!]
    \centering
    \begin{subfigure}[b]{0.24\textwidth}
         \centering
         \includegraphics[width=\textwidth]{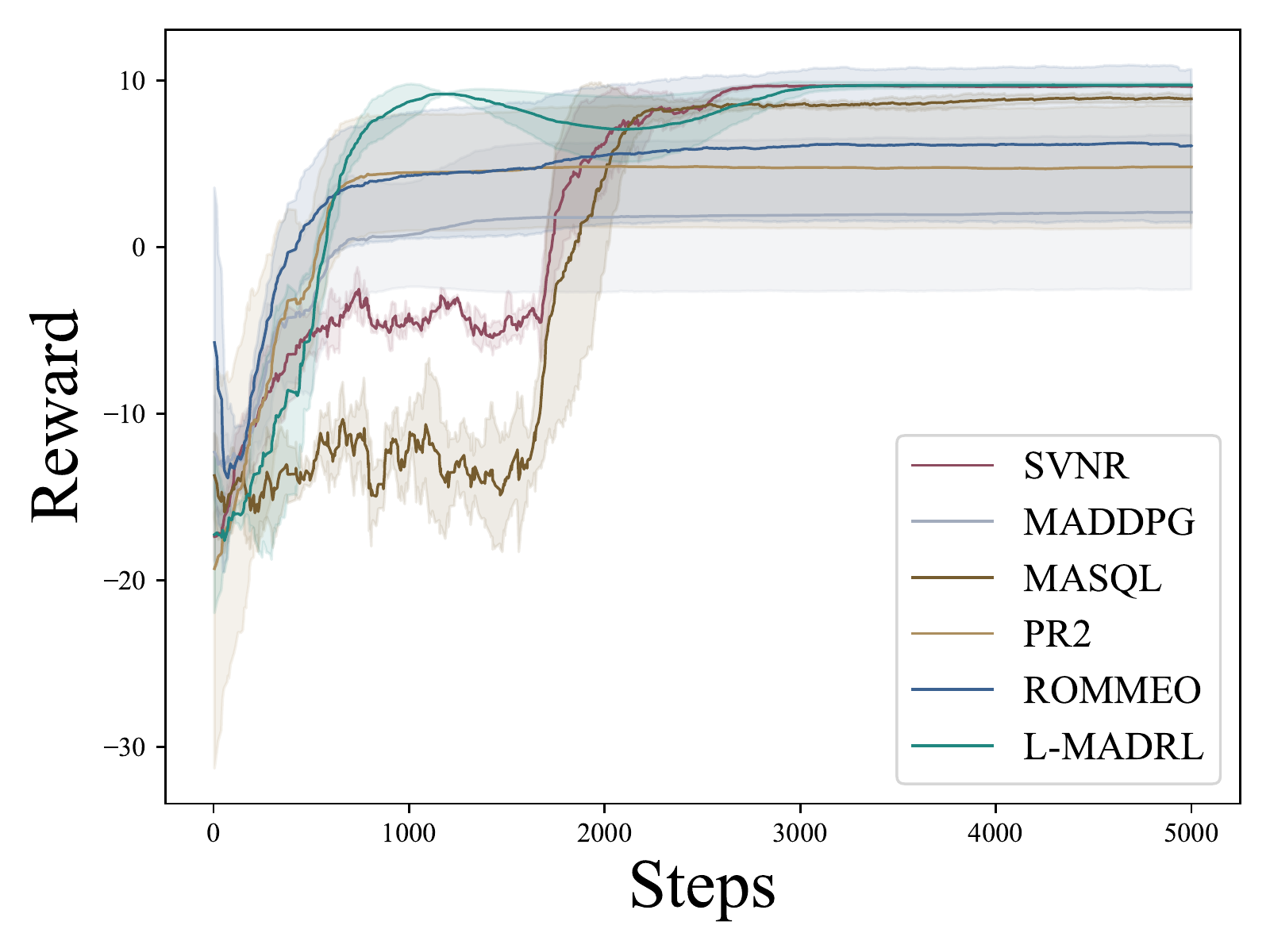}
         \caption{$s_2=3.0$}
         \label{fig: diff30}
    \end{subfigure}
    \begin{subfigure}[b]{0.24\textwidth}
         \centering
         \includegraphics[width=\textwidth]{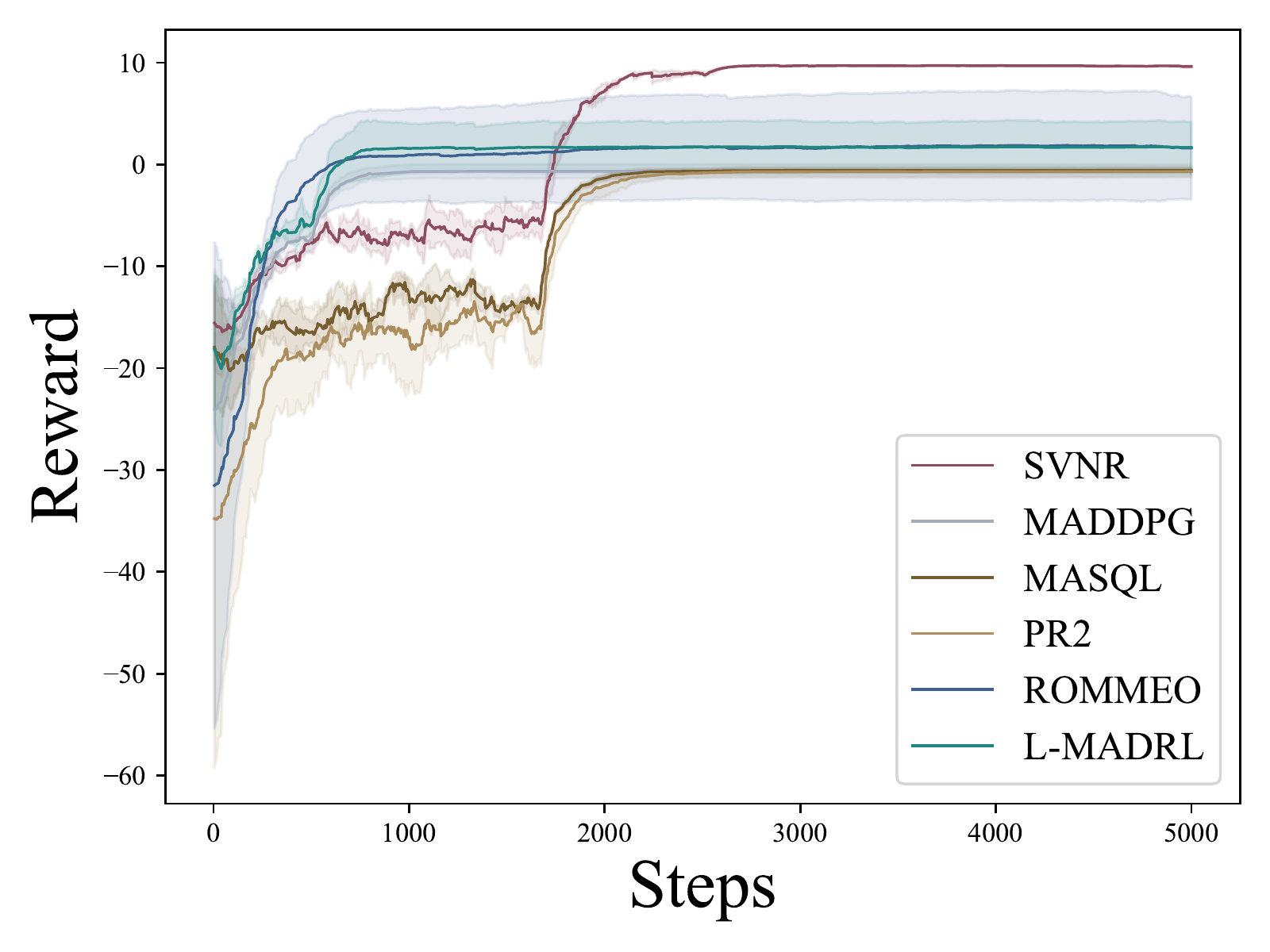}
         \caption{$s_2=2.0$}
         \label{fig: diff20}
    \end{subfigure}
    \begin{subfigure}[b]{0.24\textwidth}
         \centering
         \includegraphics[width=\textwidth]{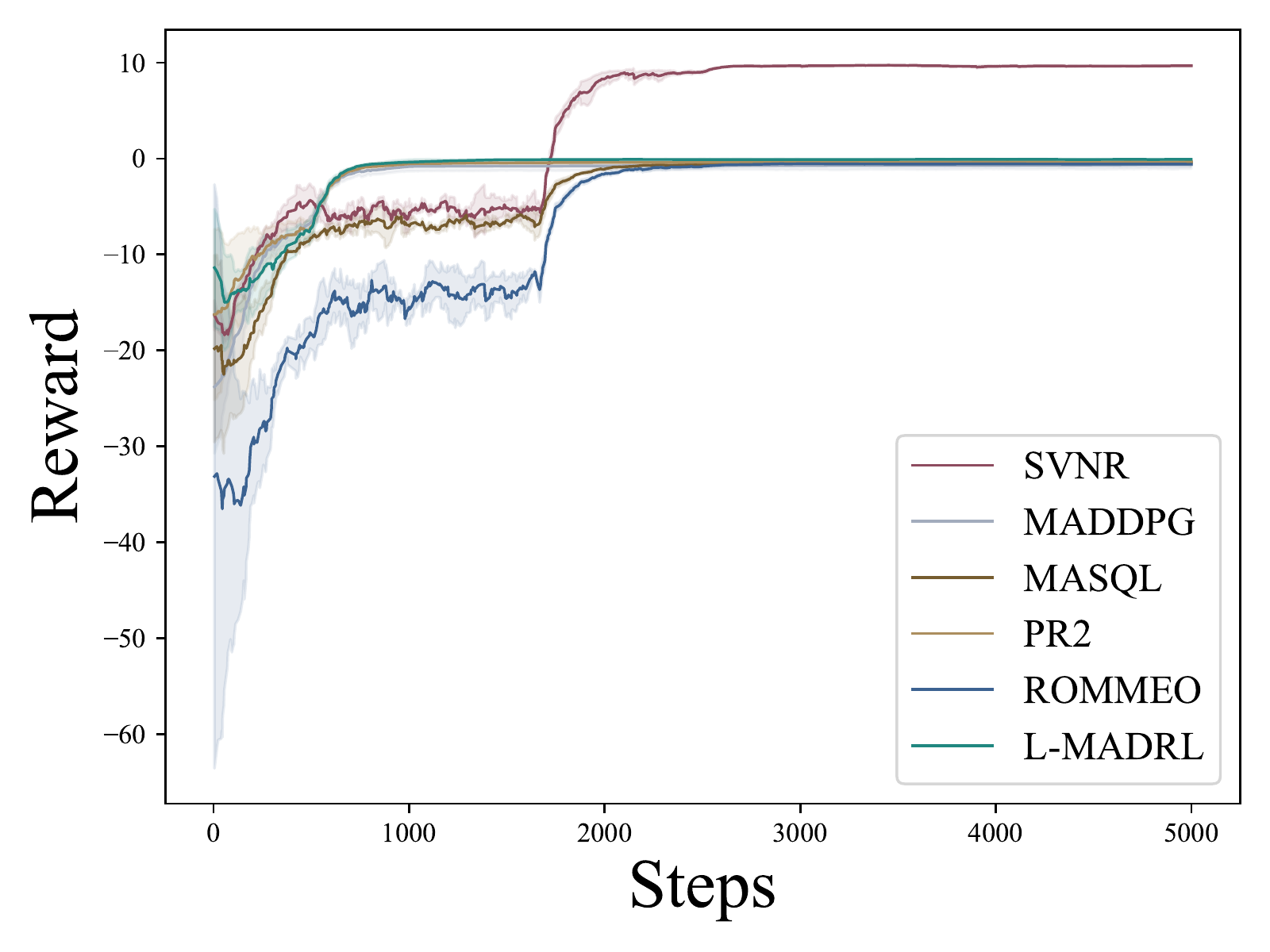}
         \caption{$s_2=1.5$}
         \label{fig: diff15}
    \end{subfigure}
    \begin{subfigure}[b]{0.24\textwidth}
         \centering
         \includegraphics[width=\textwidth]{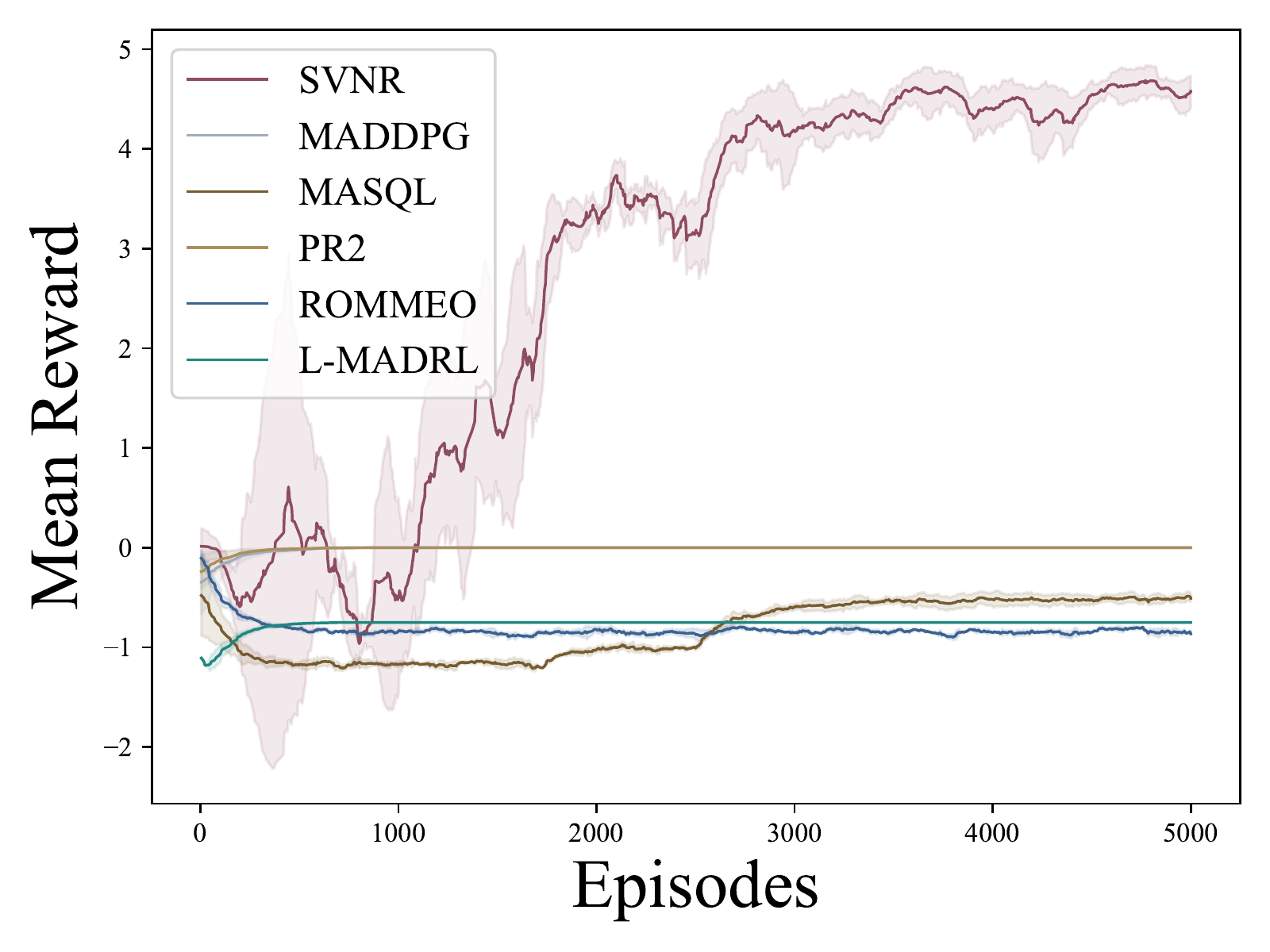}
         \caption{Particle Gather}
         \label{fig: lc-gather}
    \end{subfigure}
    
    \caption{Influence of different coverage factors $s_2$ on the training curves of (a-c) our method and different baselines in the \textit{Max Of Three}. (d) shows the training curves in the \textit{Particle Gather} scenario. The solid lines and shadow areas denote the mean and variance of the instantaneous rewards with $5$ different seeds. With the larger $s_2$, the agents encounter a higher impact of \textit{relative over-generalization}, and the proposed SVNR achieves the optimal solution in all settings.}
    \label{fig: diff-res}
\end{figure*}

\begin{figure*}[htb!]
    \centering
    \begin{subfigure}[b]{0.32\textwidth}
         \centering
         \includegraphics[width=\textwidth]{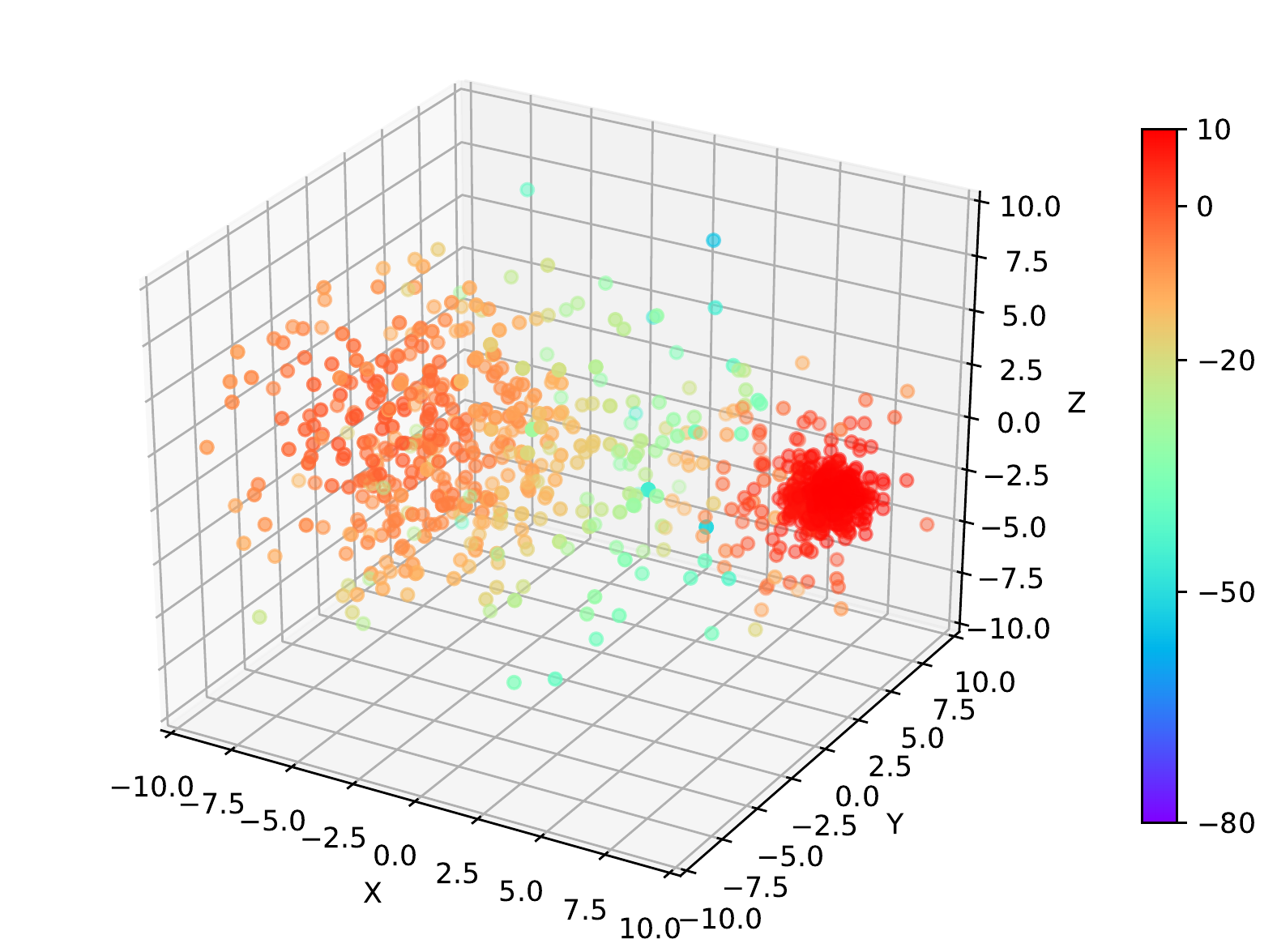}
         \caption{SVNR~(Ours)}
         \label{fig: diff-ACF}
    \end{subfigure}
    \begin{subfigure}[b]{0.32\textwidth}
         \centering
         \includegraphics[width=\textwidth]{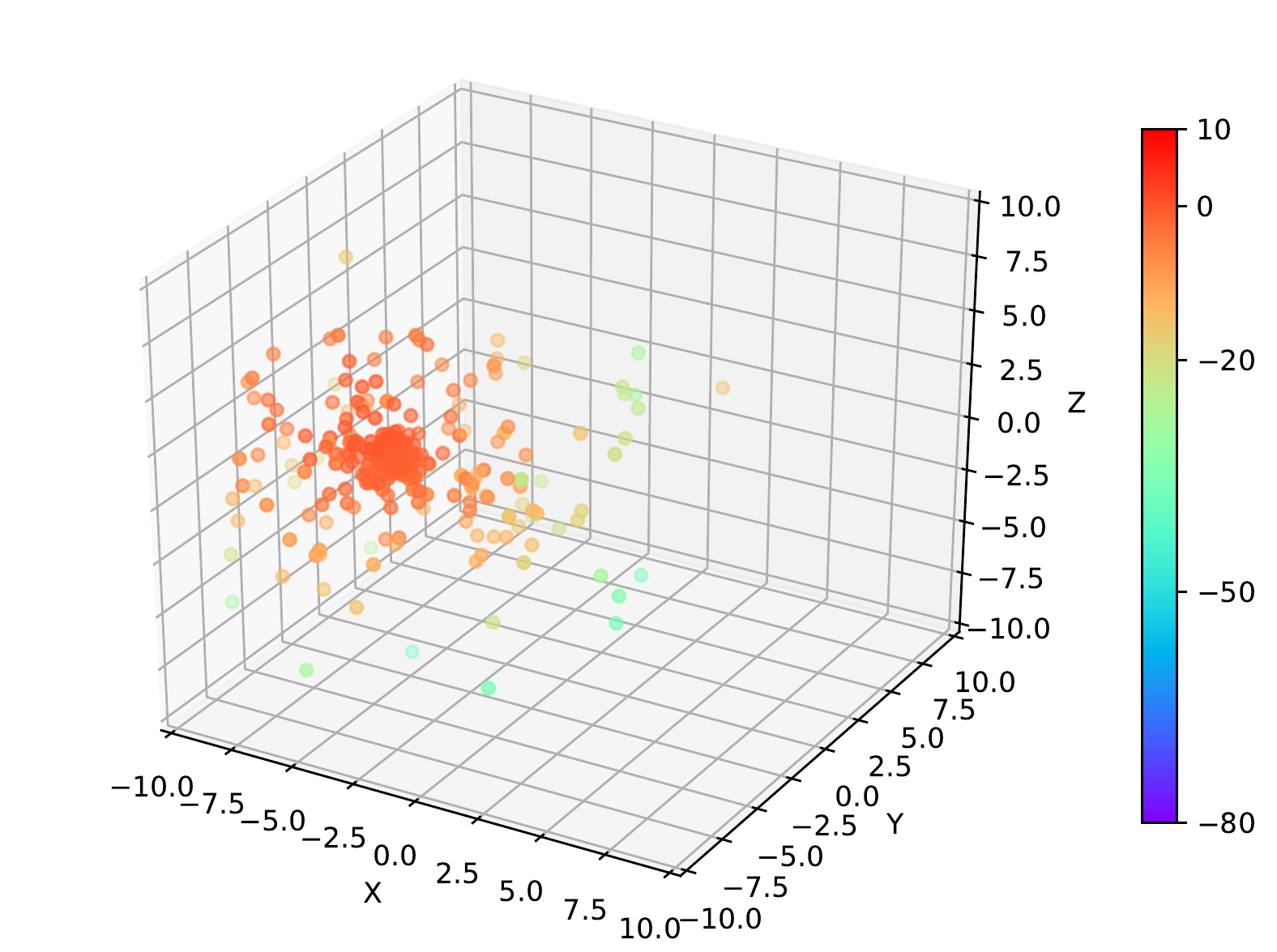}
         \caption{MADDPG}
         \label{fig: diff-maddpg}
    \end{subfigure}
    \begin{subfigure}[b]{0.32\textwidth}
         \centering
         \includegraphics[width=\textwidth]{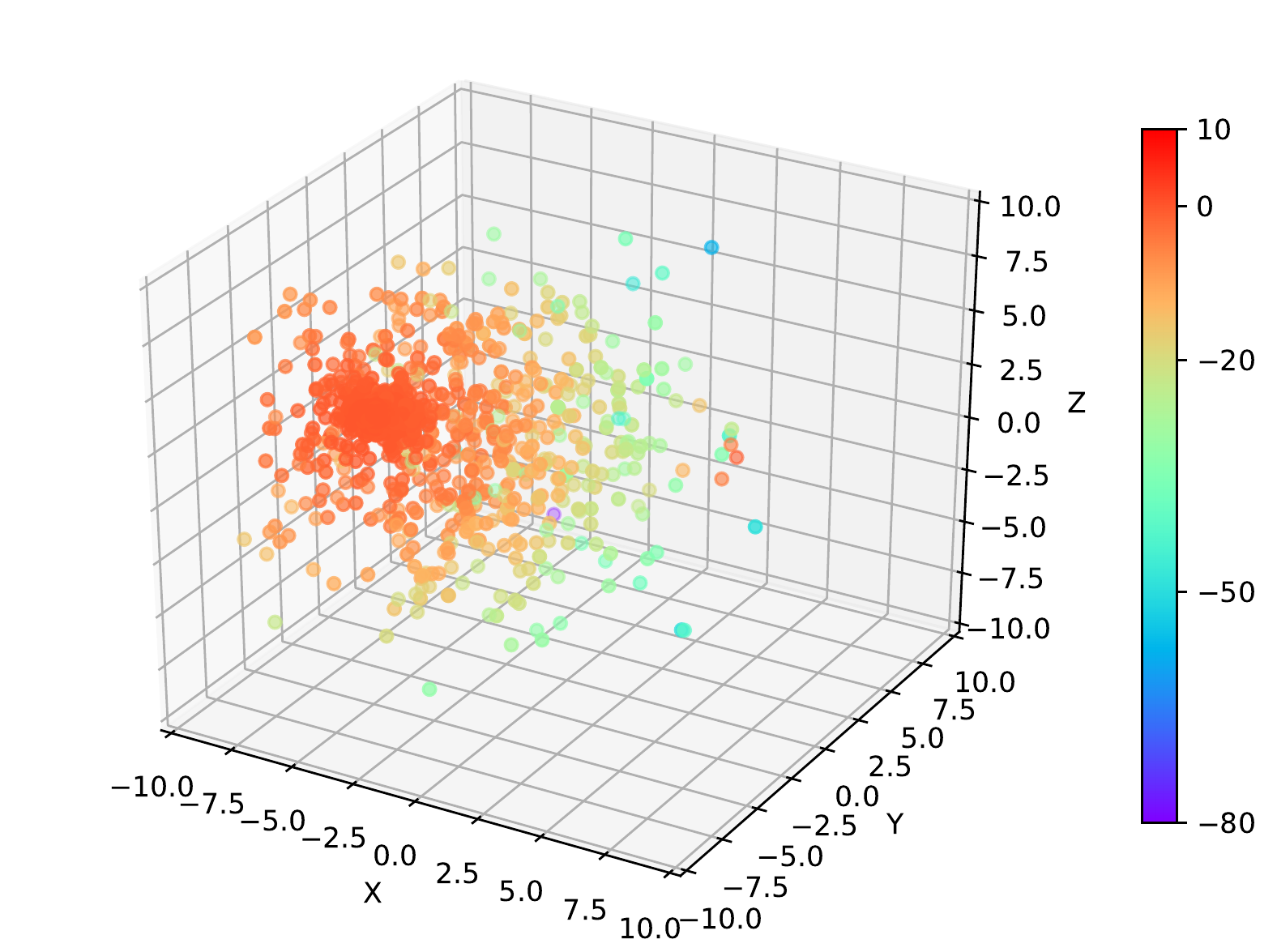}
         \caption{MASQL}
    \end{subfigure}
    
    \begin{subfigure}[b]{0.32\textwidth}
         \centering
         \includegraphics[width=\textwidth]{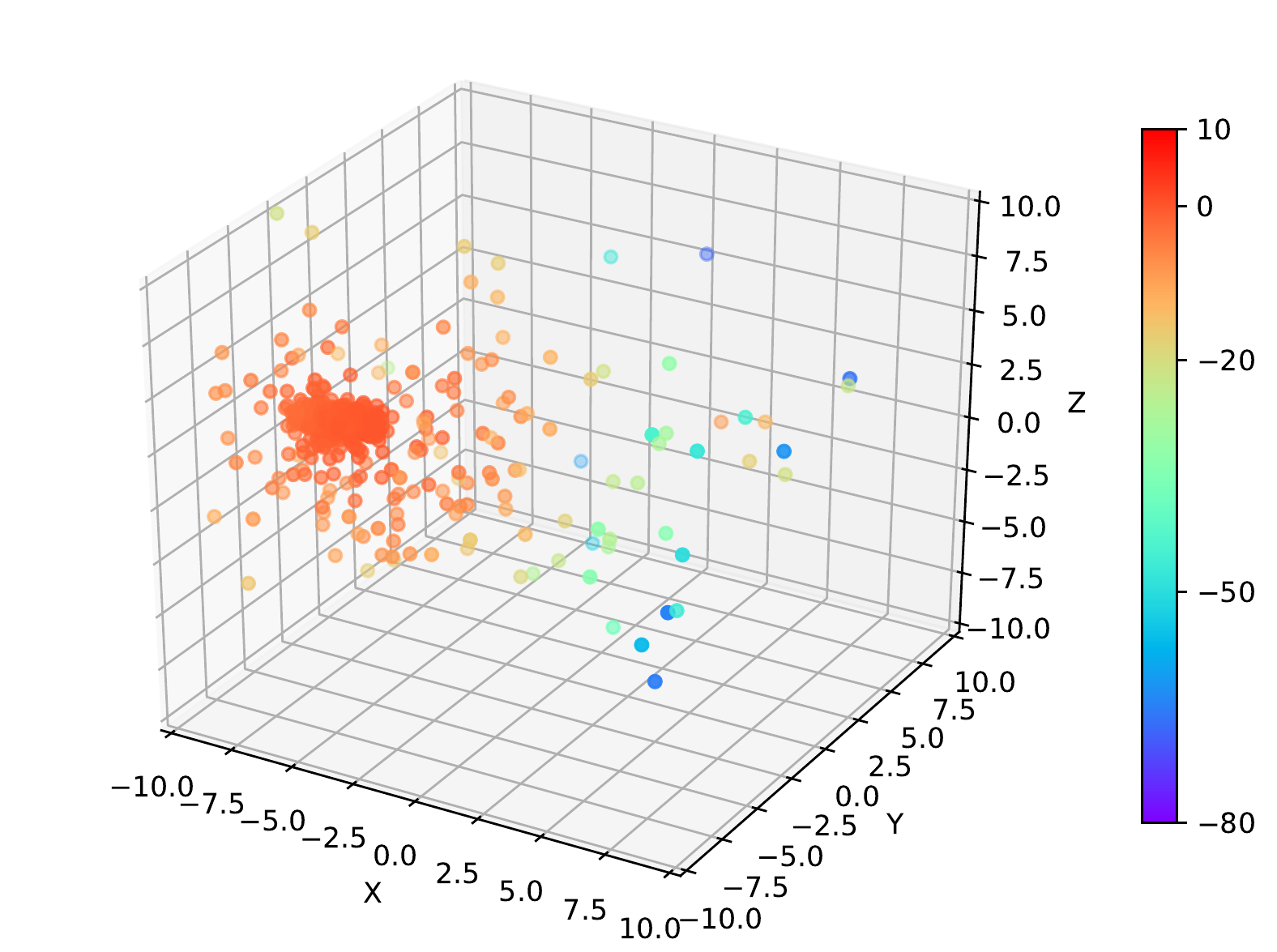}
         \caption{PR2}
         \label{fig: diff-pr2}
    \end{subfigure}
    \begin{subfigure}[b]{0.32\textwidth}
         \centering
         \includegraphics[width=\textwidth]{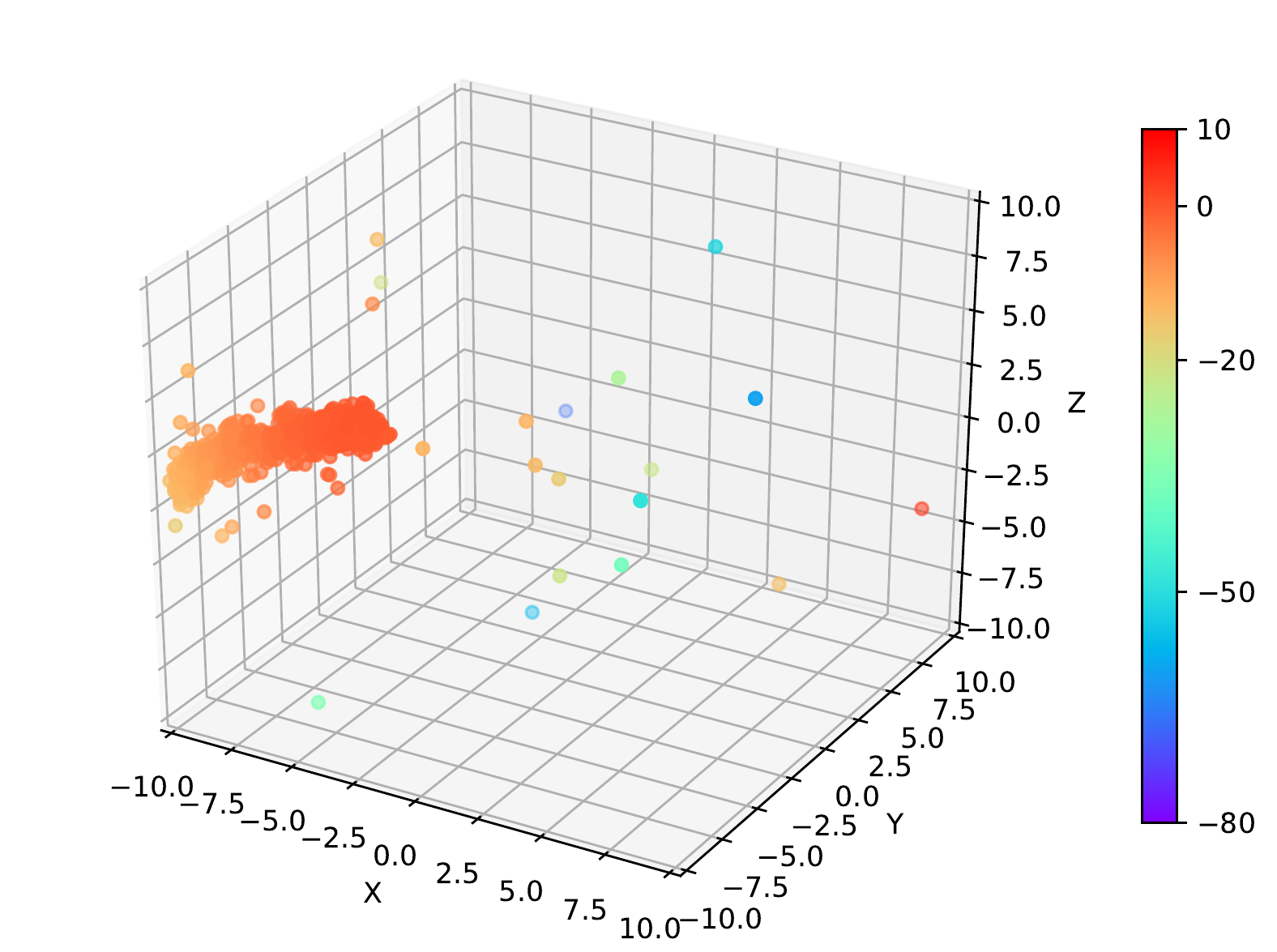}
         \caption{ROMMEO}
         \label{fig: diff-rommeo}
    \end{subfigure}
    \begin{subfigure}[b]{0.32\textwidth}
         \centering
         \includegraphics[width=\textwidth]{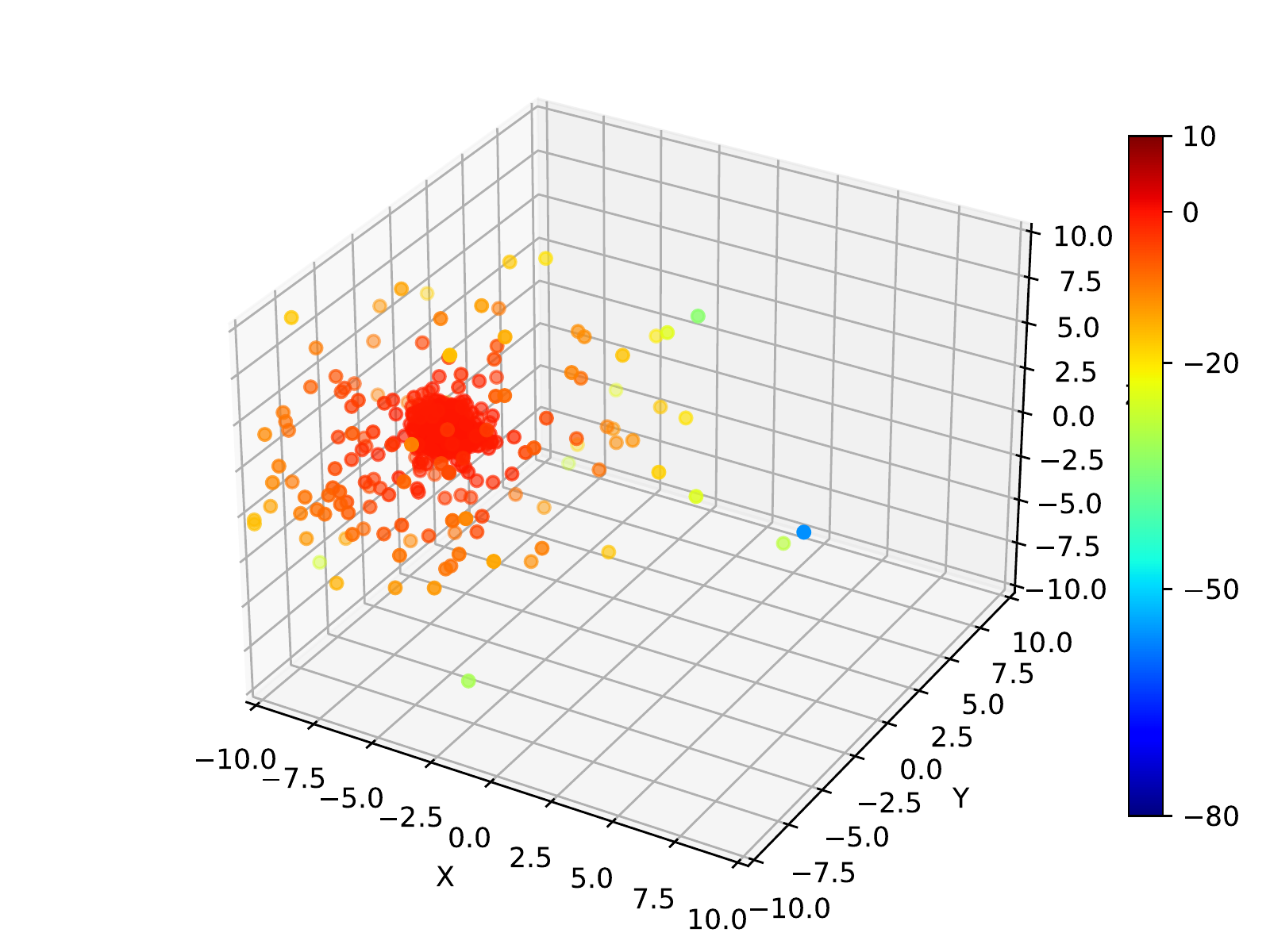}
         \caption{L-MADRL}
    \end{subfigure}
    \caption{The sampled joint actions of (a) our method and (b-d) some representative baselines under the settings of Figure~(\ref{fig: diff15}) from $1$ to $3000$ training timesteps. Each point represents a joint action taken by the agents at the corresponding timestep, and different colors represent the levels of instantaneous rewards.}
    \label{fig: diff-dyna}
\end{figure*}

Although the formulation of the game is relatively simple, it poses great difficulty to gradient-based algorithms as in almost all the joint action space. The gradient points to a sub-optimal solution.
As shown in Figure~\ref{fig: diff-res}, the MADDPG algorithm falls into the local optimum ({\em{i.e.}}, the reward is $0$) under all settings. 
MASQL, PR2, and ROMMEO can only jump out of the local optimum under the relatively simple setting ({\em{i.e.}}, $s_2=3.0$), and they all have significant variance.
Our method, SVNR, can steadily converge to the global optimum while jumping out of the local optimum under all settings.
Moreover, we test the execution performances for all the methods as shown in Table~\ref{tb: dec-result}. Our SVNR also achieves the highest return.
This shows the superiority of SVNR in addressing RO.

To better understand the learning behavior in the {\sc{Max of Three}}, we visualize the learning dynamic of SVNR and some other baselines under $s_2=1.5$ in Figure~\ref{fig: diff-dyna}.
Each point represents a joint action taken by the agents from $1$ to $3000$ steps. 
Different colors represent the levels of instantaneous rewards.
During $1$ to $1500$ steps, SVNR agents have a significant visitation probability on the local optima (the left side at Figure~\ref{fig: diff-ACF}).
They visit the global optima more frequently at $1500$ to $3000$ steps while exploring the other area.
With the learning process kept on, SVNR converges to the $10$ step reward as shown in Figure~\ref{fig: diff15}.
Other baselines are concentrated near the local optimum.

\noindent{\sc{PRO-Challenged}}. We set $h_2=1.0$, $s_2=2$, $x_2=7$, $y_2=7$, $z_2=-3$, $c_1=c_2=10$ in the differential game to construct the \textit{Two Modalities} scenario as the PRO-Challenged scenario.
There exists two points $(-5, -5, 3)$ and $(7,7,-3)$ that have the highest, $10$, utility.
Thus the optimal perceived joint policy should capture the two modalities. However, when agents do not know the optimal opponent policy, they usually tend to converge to one single modality, and PRO happens.
We train each method with $5000$ episodes and visualize their converged perceived joint policies by sampling.
As shown in Figure~\ref{fig: two-modality}, our SVNR captures the two modalities of the game while other baselines converge to the single modality policy.

\subsection{Particle Gather}\label{sec:rep-game}

The \textit{Particle Gather} is built with \textit{Multi-Agent Particle World}~\cite{maddpg}.
There are two particles in a continuous physical world.
Each particle is controlled by two agents, the $x$-agent and the $y$-agent, which control the particle's movement together.
When two particles reach a fixed landmark, four agents are rewarded with $5$ together; 
Moreover, if only one particle reaches the landmark, all the agents are penalized by $-2$.
Otherwise, there is no instantaneous reward ({\em{i.e.}}, four agents are rewarded by $0$) that will be feedback to all agents.
This iterated continuous game lasts for $25$ timesteps. 
The goal of all agents is to maximize the individual expected cumulative reward for $25$ timesteps.
This scenario is difficult because without knowing others' actions, the best choice for all the agents will be to get far away from the landmark, making the optimal policy (reach the landmark simultaneously) hard to obtain.

All methods are trained for $5000$ episodes, which consists of $25$ timesteps, with tuned hyperparameters, and the learning curves are shown in Figure~\ref{fig: lc-gather}.
It shows that all baselines converge to the worst solution except for PR2 and MADDPG falling into the local optimum.
Our method still steadily converges to the global optimum while jumping out of the local optimum.
Table~\ref{tb: dec-result} also shows the highest test performance of our SVNR compared with baselines.

\subsection{Ablation Studies}

We take ablation studies on the $C$'s design.
There are two typical $C\in \mathbb{C}_{\mathrm{Nested}}$, {\em{i.e.}}, full negotiation and strict nested negotiation.
Our SVNR adopts the nested decomposition that $C_i = \{1, \dots, i\}$.
We design SVNR-F, which adopts $C_i = -i$ to show whether making conditions on more agents can improve the performance.
Moreover, we also devise SVNR-M as another baseline which is the proposed SVNR adopt $C_i=\{\}$.
This can be useful to show the importance of let $C_i\in \mathbb{C}_\mathrm{Nested}$.
We take the experiments on the \textit{Max of Three} and \textit{Particle Gather} for further analysis.


\begin{figure}[htb!]
    \centering
    \begin{subfigure}[b]{0.24\textwidth}
         \centering
          \includegraphics[width=\textwidth]{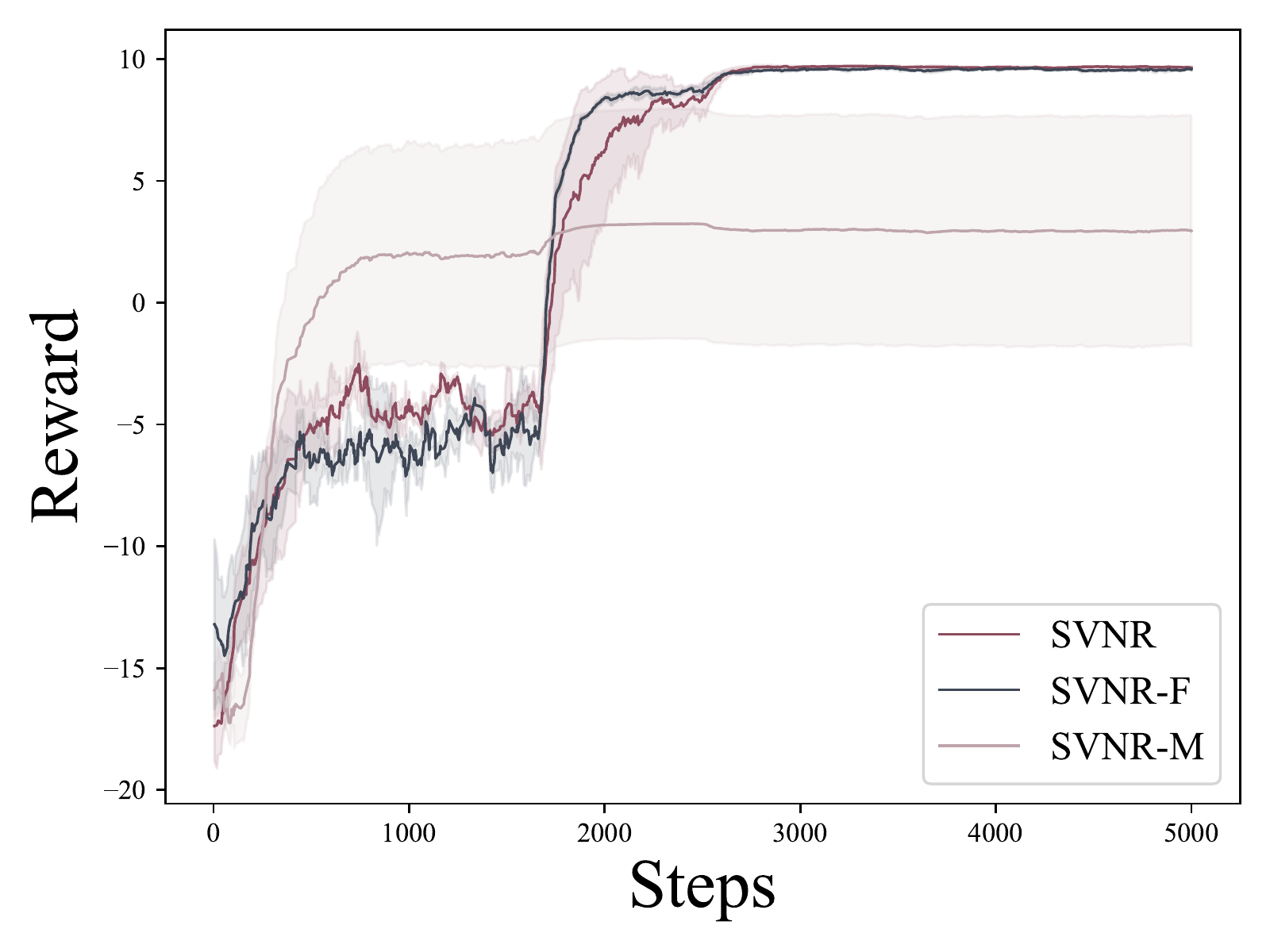}
         \caption{$s_2=3.0$}
    \end{subfigure}
    \begin{subfigure}[b]{0.24\textwidth}
         \centering
         \includegraphics[width=\textwidth]{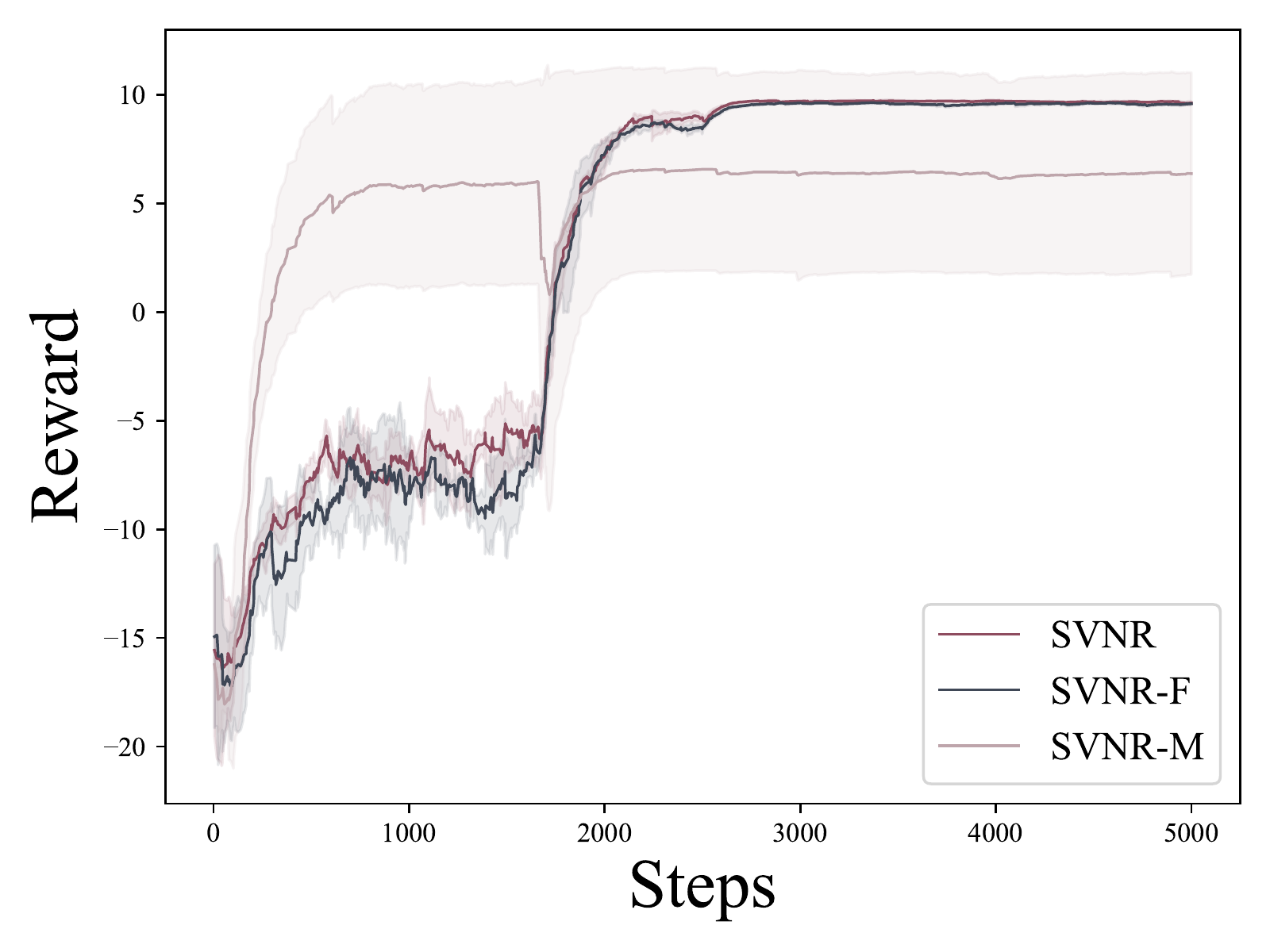}
         \caption{$s_2=2.0$}
    \end{subfigure}
    \begin{subfigure}[b]{0.24\textwidth}
         \centering
         \includegraphics[width=\textwidth]{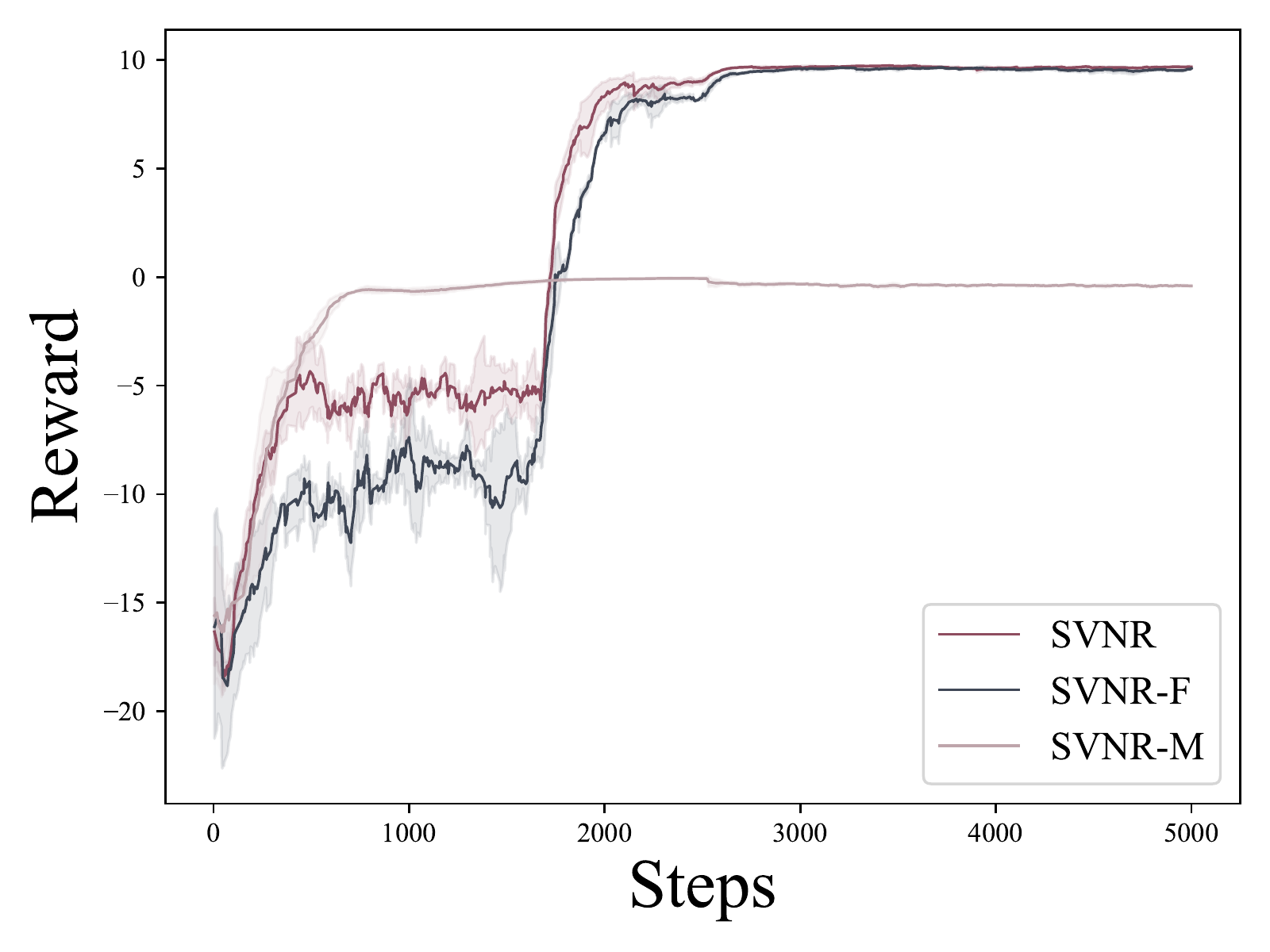}
         \caption{$s_2=1.5$}
    \end{subfigure}
    \begin{subfigure}[b]{0.24\textwidth}
         \centering
         \includegraphics[width=\textwidth]{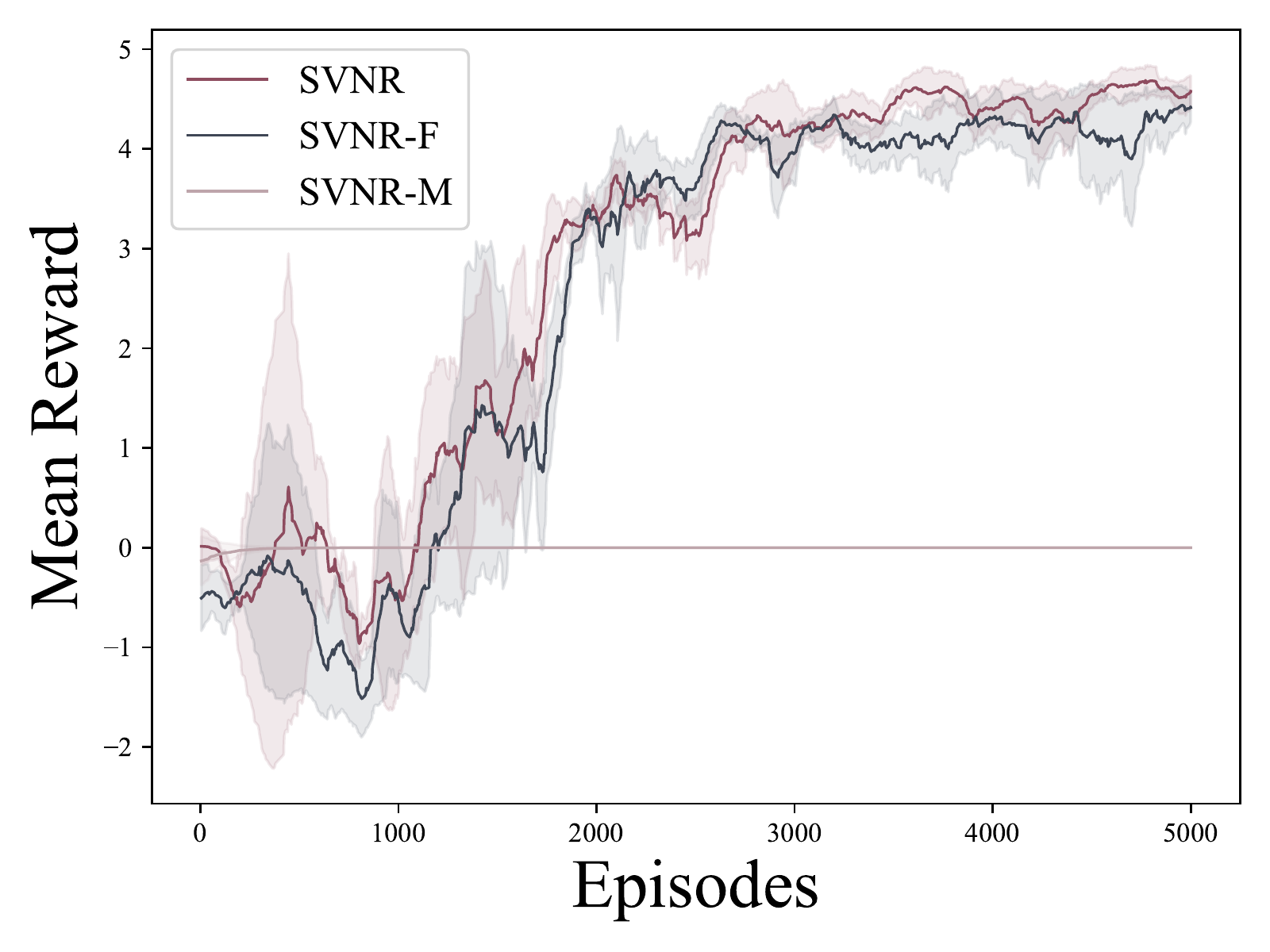}
         \caption{Particle Gather}
         \label{appfig: gather}
    \end{subfigure}
    \caption{Influence of different coverage factors $s_2$ on the training curves of (a-c) our method and different baselines in the \textit{Max Of Three}. (d) shows the training curves in the \textit{Particle Gather} scenario. The solid lines and shadow areas denote the mean and variance of the instantaneous rewards with $5$ different seeds. With the larger $s_2$, the agents encounter a higher impact of \textit{relative over-generalization}, and the proposed SVNR achieves the optimal solution in all settings.}
    \label{appfig: ablation}
\end{figure}

\begin{figure}[htb!]
    \centering
    \begin{subfigure}[b]{0.32\textwidth}
         \centering
         \includegraphics[width=\textwidth]{./diff-ncf}
         \caption{SVNR~(Ours)}
    \end{subfigure}
    \begin{subfigure}[b]{0.32\textwidth}
         \centering
         \includegraphics[width=\textwidth]{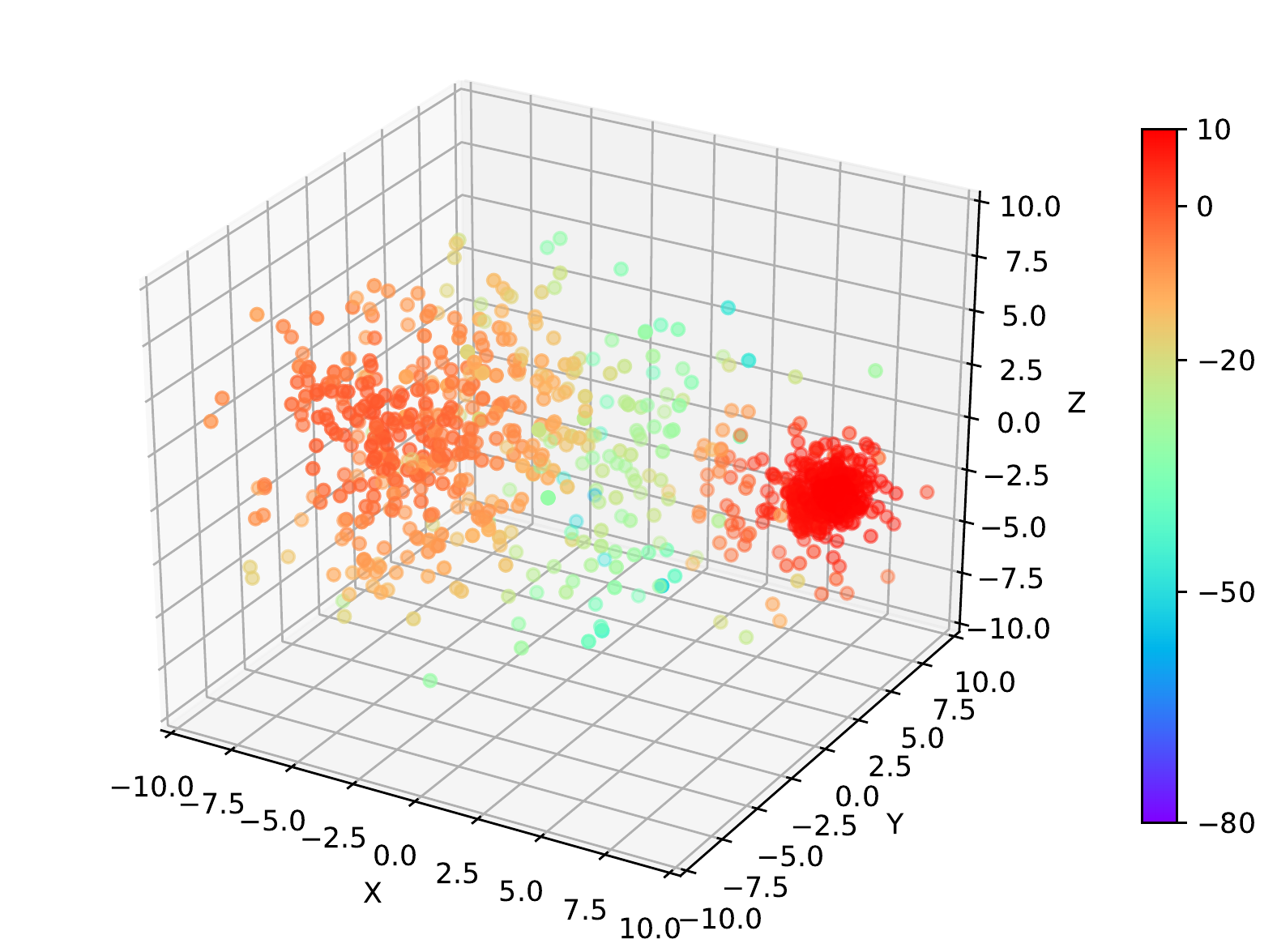}
         \caption{SVNR-F}
    \end{subfigure}
    \begin{subfigure}[b]{0.32\textwidth}
         \centering
         \includegraphics[width=\textwidth]{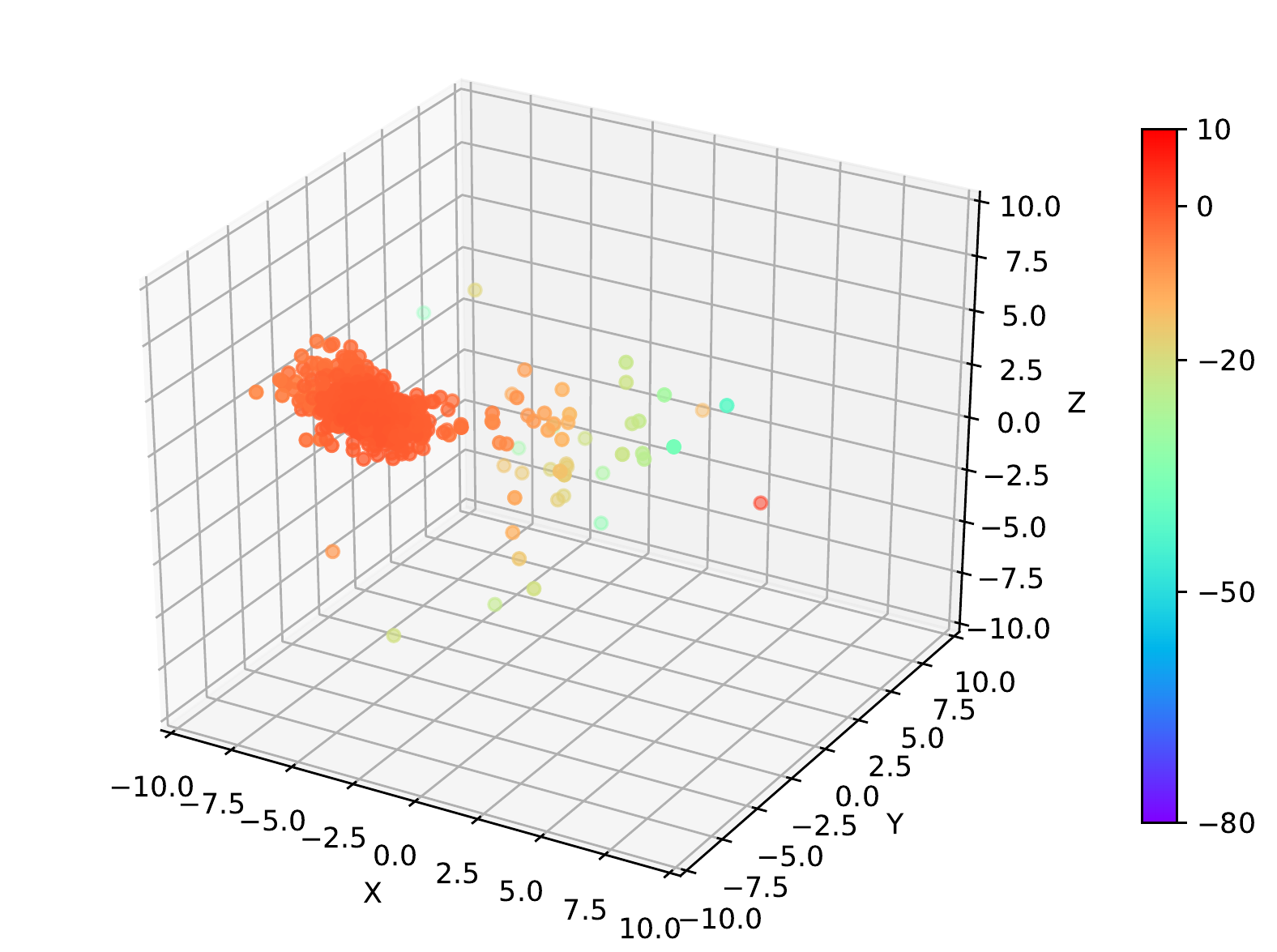}
         \caption{SVNR-M}
    \end{subfigure}

    \caption{The $1000$ sampled joint actions of all methods in the setting of $s_2=1.5$ in the \textit{Max of Three} scenario. Each point represents a joint action taken by the agents at a specific timestep, and different colors represent the levels of instantaneous rewards. All joint actions are sampled every $3$ timestep from $1$ to $3000$ timesteps in the training phase.}
    \label{appfig: vis}
\end{figure}

As shown in Figure~\ref{appfig: ablation}, both the SVNR and SVNR-F outperform the SVNR-M under $s_2=1.5, 2.0, 3.0$ in the \textit{Max of Three} scenario, which indicates the necessity of taking other agents' noises into consideration.
We also visualize their joint actions from $1$ to $3000$ steps under $s_2=1.5$ as shown in Figure~\ref{appfig: vis}.
Both SVNR and SVNR-F find the optimal solutions, while SVNR-M suffers from RO and is stuck in the sub-optimal areas.

Experiments on \textit{Particle Gather} show similar results to those shown in Figure~\ref{appfig: gather}.
As shown in the figure, both the SVNR and SVNR-F outperform the SVNR-M in the \textit{Particle Gather} scenario, which indicates the necessity of taking other agents' noises into consideration again.
\section{Conclusion and Discussions}\label{sec: conc}
This paper proposes a novel reasoning framework, Negotiated Reasoning (NR), for MARL and establishes theoretical connections between Relative Over-generalization and NR.
Then we derive an RO-free negotiated reasoning, Stein Variational Negotiated Reasoning (SVNR), which is derived based on stein variational gradient descent.
By integrating SVNR to maximum entropy policy iteration, it is provably RO-free and converges to optimal cooperation. 
Further, we propose a practical implementation with neural network approximation.
Empirically, SVNR also outperforms baselines in addressing RO and approaching optimal cooperation.

There are also some limitations of our method. One main limitation of our SVNR approach is its scalability in scenarios involving many agents. Both SVNR and other existing reasoning-based MARL methods require at least one agent to perform reasoning on all other agents. As the number of agents increases, it becomes increasingly challenging to accurately reason and make the best response.
To address this scalability issue, potential solutions include the use of attention mechanisms or domain knowledge that can help generate sparse negotiation structures. These represent interesting and promising directions for future research in this area. Moreover, investigating the theoretical and practical gap will be interesting when we sparsify the negotiation structure which may violate the nested negotiation structure requirement.
We leave them as future works. 
\bibliographystyle{plain}  
\bibliography{references}  

\appendix
\onecolumn

\section{Theorems}
\begin{theorem}[Nested factorization requirement]\label{th:nested}
For a policy factorization method that adopts local policies $\{\pi_1(u_1\mid \boldsymbol{u}_{C_1}), \cdots, \pi_N(u_N\mid \boldsymbol{u}_{C_N})\}$ to represent the joint policy $\pi_{\mathrm{jt}}(\boldsymbol{u})$, it can achieve full joint policy representation capacity if and only if there exists a permutation $\sigma$ of $[N]$ that satisfies
$$\{ i+1, \cdots,N \}\subset \left\{ \sigma(j)\ |\ \forall \ j\in C_{\sigma^{-1}(i)} \right\},\quad \forall i.$$
For simplicity we denote as $\mathbf{C}= \{C_1,\dots,C_N\} \in \mathbb{C}_{\mathrm{Nested}}$ and the $ \mathbb{C}_{\mathrm{Nested}}$ is called Nested Coordination Space.
\end{theorem}The proof of Theorem \ref{th:nested} can be found in Appendix~\ref{proof: the3.1}.
The above theorem urges us to decompose the joint policy into conditional policies that satisfy the nested requirement.
ROMMEO takes $C_i = -i,$ $ \forall i$, which satisfies our nested factorization requirement and achieves the full capacity.

\section{Derivations}
\label{sec: der}

\renewcommand*{\proofname}{Derivation}

\subsection{Derivation of Equation 12}
\label{der: phi*}
\begin{proof}
As proved in the MPSVGD~\cite{mpsvgd}, for a graphical model $p(\boldsymbol{z})\propto \prod_{i=1}^N p(z_i\mid \boldsymbol{z}_{C_i})$, let $\mathbf{z}={T}(\mathbf{x})=\left[x_{1}, \ldots, T_{i}\left(x_{i}\right), \ldots, x_{N}\right]^{\top}$ with $T_{i}: x_{i} \rightarrow x_{i}+\epsilon \phi_{i}\left(\mathbf{x}\right), \phi_{i} \in \mathcal{H}_{i}$ where $\mathcal{H}_{i}$ is a Reproducing kernel Hilbert Space~(RKHS)
associated with the local kernel $k_{i}: \mathcal{X} \times \mathcal{X} \rightarrow \mathbb{R}$, we have
\[
\begin{array}{l}
\nabla_{\epsilon} \mathrm{KL}\left(q_{[T]} \| p\right)= 
\nabla_{\epsilon} \mathrm{KL}\left(q_{\left[T_{i}\right]}\left(z_{i} \mid \mathbf{z}_{C_{i}}\right) q\left(\mathbf{z}_{C_{i}}\right) \| p\left(z_{i} \mid \mathbf{z}_{C_{i}}\right) q\left(\mathbf{z}_{C_{i}}\right)\right),
\end{array}
\]
and the solution for $\left.\min _{\left\|\phi_{i}\right\|_{\mathcal{H}_{i}} \leq 1} \nabla_{\epsilon} \mathrm{KL}\left(q_{[\boldsymbol{T}]} \| p\right)\right|_{\epsilon=0}$ is
$\phi_{i}^{*} /\left\|\phi_{i}^{*}\right\|_{\mathcal{H}_{i}}$, where
\[
\phi_{i}^{*}(\mathbf{x})= \mathbb{E}_{\mathbf{y} \sim q}[k_{i}(\mathbf{x}_{C_i}, \mathbf{y}_{C_i}) \nabla_{y_{i}} \log p(y_{i} \mid \mathbf{y}_{C_i}) +\nabla_{y_{i}} k_{i}(\mathbf{x}_{C_i}, \mathbf{y}_{C_i})].
\]
Under mild conditions as states in the MPSVGD~\cite{mpsvgd}, the convergence condition $\phi_{i}^{*}(\mathbf{x}) = 0 $ if and only if $q(x_i|\boldsymbol{x}_{C_i})=p(x_i|\boldsymbol{x}_{C_i})$.
Take $p^\phi$ and $\exp{(Q^\theta)}$ as $q$ and $p$ respectively, then
\begin{equation}
\begin{aligned}
&\Delta f^{\boldsymbol{\phi}}_i \left(\cdot ; s_{t}\right)= \mathbb{E}_{\boldsymbol{u}\sim p^{\phi}} \left[ \kappa_i(\boldsymbol{u}_{S_i}, p^{\boldsymbol{\phi}}_{S_i}(\cdot ; {s}_{t})) \nabla_{u^{\prime}_i} Q^{\theta}\left({s}_{t}, \boldsymbol{u}^{\prime}\right)\big|_{\boldsymbol{u}^{\prime}=\boldsymbol{u}}\right. \left.  +\, \alpha_i \nabla_{\boldsymbol{u}^{\prime}_i} \kappa_i(\boldsymbol{u}_{S_i}^{\prime}, p^{\boldsymbol{\phi}}_{S_i}(\cdot ; {s}_{t})) \big|_{\boldsymbol{u}^{\prime}=\boldsymbol{u}} \right],
\end{aligned}
\end{equation}
where $S_i:= \{i\}\bigcup C_i$.
\end{proof}

\subsection{Derivation of Equation~9}
\begin{proof}
One direct way to update the parameter $\phi_i$ is to obtain $z$ by running MPSVGD until convergence and update $\phi_i$
\[
\phi_i^{t+1} \leftarrow \argmin_{\phi_i} \sum_{k=1}^K \|p^{\phi^t}(\xi^k;s) - z^k \|_2^2.
\]
To gain a more computationally efficient approximation, we perform one gradient descent step
\[
\phi_i^{t+1} \leftarrow \phi_i^{t} + \epsilon\cdot\mathbb{E}_{\xi}\left[\Delta f^{\phi^t}_i \left(\xi ; s_{t}\right) \frac{\partial f^{\phi^t}_i\left(\xi; s_{t}\right)}{\partial \phi_i}\right],
\]
with a small step size $\epsilon$.
\end{proof}


\section{Proofs}

\renewcommand*{\proofname}{Proof}

\subsection{Proof for Theorem~\ref{the: pro-free-nr}}
\label{proof: pro-free-nr}
In NR framework, each agent $i$ holds $\hat{\pi}_i=p(\boldsymbol{u}^k\mid s)$, 
if $\lim_{k\rightarrow K}p(\boldsymbol{u}^k\mid s) \rightarrow \pi^*(\boldsymbol{u}^k \mid s)$, 
then 
$$
\min_{f_i}D_{KL} (f_i p(\boldsymbol{u}^K_{-i}\mid s) \| \pi^*_\alpha) = \min_{f_i}D_{KL} (f_i \pi^*(\boldsymbol{u}^K_{-i}\mid s)) \| \pi^*_\alpha).
$$
Thus it is PRO-free after $K$ reasoning rounds. 

\subsection{Proof for Theorem~\ref{the: ero-free-nr}}
\label{proof: ero-free-nr}
If $\alpha \rightarrow 0$, then $\pi^*_\alpha$ approaches to the maximum utility 
$$
U^{\pi^*_\alpha} = \max_{\pi} U^\pi, \quad \alpha\rightarrow 0,
$$
due to $Q_{\mathrm{soft}} = U^{\pi^*_\alpha}+\sum_t \mathbb{E}_{(s_t, \boldsymbol{u}_t)\sim \beta_{\pi^*}} H(\pi^*(\cdot\mid s_t))$. 
For PRO-free agents in NR, $p(\boldsymbol{u}^K\mid s) = \pi^*_\alpha(\boldsymbol{u}^k \mid s)$ and $\alpha\rightarrow 0$, take $\bar{\pi}_i=u_i^{0,K}$, then 
$$
\max_{\pi_i} U^{\pi_i\prod_{j\neq i} \pi_j^\prime} = U^{\bar{\pi}^\prime}.
$$ 
Thus they are ERO-free. 
\subsection{Proof for Theorem~\ref{th:nested}}
\label{proof: the3.1}

\begin{proof}
The conditional theorem~\cite{cond2joint} proves that the $\{\pi_1(u_1\mid \boldsymbol{u}_{C_1}), \dots, \pi_N(u_N\mid \boldsymbol{u}_{C_N})\}$ uniquely determines the joint policy if and only if the $\mathbf{C}\in \mathbb{C}_{\mathrm{Nested}}$.
For any joint policy $\pi$, we can obtain $$\pi_i(u_i\mid \boldsymbol{u}_{C_i}) = \frac{\int \pi(\boldsymbol{u})d {\boldsymbol{u}_{\{i\}\bigcup C_i}}}{\int \pi(\boldsymbol{u})d {\boldsymbol{u}_{C_i}}}, \quad \forall 1 \leq i \leq N .$$
When the $\mathbf{C}\in \mathbb{C}_{\mathrm{Nested}}$, the conditional policies uniquely determine the joint policy.
Then for arbitrary joint policy, we can represent it as the nested conditional policies, and Theorem~\ref{th:nested} gets proved.
\end{proof}

\subsection{Proof for ERO-free property of SVNR}
\label{proof: nr-converge}
We first prove the strict nested negotiation makes SVNR converge (i.e., the first condition in \eqref{eq: cond-nego}). 
Without loss of generalization, we take $C_i=\{1, \dots, i\}$ for every agent $i$.
For agent $1$, $C_1=\{1\}$ and the \eqref{eq: par-ite} degenerate to the SVGD, which has been proved weakly converged to target distribution $\pi^*(u_1)$ in \cite{svgd-theory}:
$$ 
\lim_{k\rightarrow K} f_1^k(u_1\mid s, \boldsymbol{u}_{C_1}^{l, k-1}) = u_1^{l, k-1}, \quad \forall l\leq M
$$
$$
\lim_{k\rightarrow K} p(u_1^k) = \pi^*(u^k_1\mid s),\quad \forall u^k_1\in \boldsymbol{\mathcal{U}}_1
$$
Then with agent $1$ converged, agent $2$'s update degenerate to the SVGD and converges to the target conditional distributions. 
Iteratively, we can obtain:
\begin{equation}
    \begin{aligned}
        &\lim_{k\rightarrow K} f_i^k(u_i\mid s, \boldsymbol{u}_{C_i}^{l, k-1}) = u_i^{l, k-1}, \quad \forall l\leq M, i \leq N, \\
        &\lim_{k\rightarrow K} p(u_i^k) = \pi^*(u^k_i\mid s, \boldsymbol{u}_{C_i}^{l, k-1}),\quad \forall u^k_i\in \boldsymbol{\mathcal{U}}_i. 
    \end{aligned}
    \label{eq: ite-prove}
\end{equation}
Thus we prove its convergence. 

According to Appendix~\ref{proof: the3.1}, the  (strict) nested conditional policies can be adopted to represent arbitrary joint policy and when the conditional policies uniquely determine the joint policy. 
Then with  \eqref{eq: ite-prove}, we have
$$
\lim_{k\rightarrow K} p(\boldsymbol{u}^k\mid s) = \pi^*(\boldsymbol{u}^k\mid s),\quad \forall u^k_i\in \boldsymbol{\mathcal{U}}_i, i\leq N. 
$$
and thus the SVNR is PRO-free. 

\subsection{Proof for Lemma~\ref{lemma:jointpolicy}}
\label{proof: jointpolicy}

\begin{proof}
We refer the readers to the SQL~\cite{sql}'s Appendix A.2.
\end{proof}

\subsection{Proof for Lemma~\ref{lemma:coorpolicy}}
\label{proof: coorpolicy}

\begin{proof}
Following the Proof~\ref{proof: nr-converge}, with K rounds of SVNR negotiation, 
\begin{equation}
    \begin{aligned}
        \hat{\pi}^{\prime} &= \lim_{k\rightarrow K} \frac{1}{M}\sum_{l=1}^M \delta_{\boldsymbol{u}^{l,k}}(\boldsymbol{u}),         \\
        &= \Tilde{\pi} = \exp{\frac{1}{\alpha}(Q(u_i, \boldsymbol{u}_{C_i}, s)-V(\boldsymbol{u}_{C_i},s))},
    \end{aligned}
\end{equation}

Then the policy improvement can be proved as in Appendix A.1 of \cite{sql}. 

\end{proof}
\subsection{Proof for Theorem 4.3}
\label{proof: the4.2}
With the Theorem~\ref{th:nested}, Lemma~\ref{lemma:jointpolicy} and Lemma~\ref{lemma:coorpolicy}, our convergence to the optimal joint policy can be similarly proved as the SQL\cite{sql}'s Appendix A.2.




\end{document}